\documentclass{article}
\pdfoutput = 1

\usepackage{microtype}
\usepackage{graphicx}
\usepackage{subfigure}
\usepackage{booktabs} % for professional tables

\usepackage[nohyperref,accepted]{icml2024}

\usepackage{amsmath}
\usepackage{amssymb}
\usepackage{mathtools}
\usepackage{amsthm}

\usepackage{soul}
\colorlet{shadecolor}{gray!20}
\sethlcolor{shadecolor}

\theoremstyle{plain}

\theoremstyle{definition}

\theoremstyle{remark}

\usepackage[textsize=tiny]{todonotes}

\usepackage[utf8]{inputenc} % allow utf-8 input
\usepackage[T1]{fontenc}    % use 8-bit T1 fonts
\usepackage{url}            % simple URL typesetting
\usepackage{booktabs}       % professional-quality tables
\usepackage{amsfonts}       % blackboard math symbols
\usepackage{nicefrac}       % compact symbols for 1/2, etc.
\usepackage{microtype}      % microtypography
\usepackage{xcolor}         % colors
\usepackage{microtype}
\usepackage{graphicx}
\usepackage{subfigure}
\usepackage{booktabs} % for professional tables

\usepackage{makecell}
\newcommand{\niparagraph}[1]{\vspace{1pt}\noindent\textbf{#1}}
\usepackage{times}
\definecolor{coolblack}{HTML}{001473}
\usepackage{graphicx}
\usepackage{yfonts}
\usepackage[linesnumbered,ruled,vlined,algo2e]{algorithm2e}
\DontPrintSemicolon
\SetKwComment{Comment}{\color{green!50!black}// }{}

\SetKwProg{Function}{function}{}{}

\usepackage{tcolorbox}
\tcbuselibrary{breakable,xparse,skins}
\usepackage{listings}

\PassOptionsToPackage{table}{xcolor}
\usepackage{savesym}
\usepackage{soul}
\usepackage{fontenc}
\usepackage{mathtools}
\usepackage{bbding}
\savesymbol{Cross}
\savesymbol{Square}
\savesymbol{TriangleUp}
\savesymbol{TriangleDown}
\usepackage[geometry]{ifsym}
\usepackage{textgreek}
\usepackage{wasysym}
\let\savedegree\Asterisk
\let\savedegree\leftmoon
\let\savedegree\rightmoon
\let\savedegree\fullmoon
\let\savedegree\newmoon
\let\savedegree\diameter

\let\diameter\relax
\usepackage{mathabx}

\let\diameter\savedegree
\usepackage{helvet}
\usepackage[export]{adjustbox}
\usepackage[toc]{multitoc}
\usepackage{changepage}

\usepackage{lipsum}
\usepackage{tikz}
\usetikzlibrary{calc}
\usepackage{hhline}
\usepackage{hyperref}
\usepackage{paralist}
\usepackage{epigraph}
\usepackage{xspace}
\usepackage{amsmath}
\usepackage{amssymb}
\usepackage[mathcal]{eucal}
\usepackage{multirow}
\usepackage{array}
\usepackage{booktabs}
\usepackage[normalem]{ulem}
\usepackage{colortbl}
\usepackage{url}
\usepackage{listings}
\usepackage{setspace}
\usepackage{enumerate}
\usepackage[hang,flushmargin]{footmisc}
\usepackage{fancyhdr}
\usepackage{appendix, apptools}
\usepackage[nottoc]{tocbibind}
\usepackage{pifont}
\usepackage{soul}
\usepackage{mdframed}
\usepackage{stmaryrd}
\usepackage[T1]{fontenc}
\usepackage{booktabs}
\usepackage{verbatim}
\usepackage{graphicx}
\usepackage{courier}
\usepackage{marginnote}

\usepackage{url}
\usepackage{bm}
\let\oldbibliography\thebibliography
\renewcommand{\thebibliography}[1]{\oldbibliography{#1}
\setlength{\itemsep}{0pt}}
\mdfdefinestyle{MyFrame}{%
    linecolor=black,
    outerlinewidth=2pt,
    roundcorner=0pt,
    innertopmargin=2pt,
    innerbottommargin=0pt,
    innerrightmargin=0pt,
    innerleftmargin=1pt}
\mdfdefinestyle{MyFrame2}{%
    linecolor=black,
    outerlinewidth=2pt,
    roundcorner=0pt,
    innertopmargin=0pt,
    innerbottommargin=0pt,
    innerrightmargin=0pt,
    innerleftmargin=1pt}

\definecolor{stringcolor}{rgb}{0.0,0.0,0.8}
\definecolor{emphcolor}{RGB}{32,63,117}
\definecolor{commentcolor}{rgb}{0.13,0.55,0.13}
\definecolor{kwcolor}{rgb}{0.0,0.0,0.0}
\definecolor{backcolor}{gray}{0.95}
\definecolor{commentgreen}{cmyk}{0.99998,0,1,0}
\definecolor{pythoncolor}{RGB}{168, 103, 13}
\definecolor{param}{RGB}{102,0,153}

\lstdefinestyle{pseudocdlt}{
    belowcaptionskip=1\baselineskip,
    backgroundcolor=\color{backcolor},
    aboveskip=3pt,
    belowskip=3pt,
    basicstyle=\linespread{1.0}\ttfamily\scriptsize,
    commentstyle=\color{commentcolor},
    keywordstyle=\bfseries\color{kwcolor},
    stringstyle=\color{stringcolor},
    morestring=[b]",
    keywordstyle=[2]{\bfseries\color{param}},
    morecomment=[l][\color{commentgreen}]{\#},
    breakatwhitespace=false,         
    breaklines=true,                 
    captionpos=b,                    
    keepspaces=true,                 
    showspaces=false,
    numberstyle=\linespread{1.0}\ttfamily,
    numbersep=5pt,
    numbers=left,
    showstringspaces=false,
    showtabs=false,                  
    tabsize=1,
    literate={\ \ }{{\ }}1,
    morekeywords={inp,out,loop,compute,transfer,param,cdlt, VMEM1, VMEM2,BUF1,BUF2,DRAM, alloc,OBUF,WBUF,IBUF,n1,c1,oh,kh,c,kh1,n,oh1},
    morekeywords=[2]{with,lambda,for,in,def,return,if,as,n1,c1,oh,kh,c,kh1,n,oh1},
    emphstyle={\bfseries\color{emphcolor}},
    emph={MAX, LOAD, MOVE,STORE, ALLOC, LOOP, FXP32,GEMM,RELU},
    keywords=[2]{IH,N,C,OH,KH,IC,OC,N1,C1,OH1,KH1,IC1,OC1},
    basewidth=0.5em
}

\newcommand{\PCignore}[1]{}

\newcommand{\xx}[1]{\texttt{#1}\xspace}

\def\Snospace~{\S{}}
\newcommand{\squishlist}{
 \begin{list}{$\bullet$}
  { \setlength{\itemsep}{0pt}
     \setlength{\parsep}{3pt}
     \setlength{\topsep}{3pt}
     \setlength{\partopsep}{0pt}
     \setlength{\leftmargin}{1.5em}
     \setlength{\labelwidth}{1em}
     \setlength{\labelsep}{0.5em} } }

\newcommand{\squishlisttwo}{
 \begin{list}{$\bullet$}
  { \setlength{\itemsep}{0pt}
     \setlength{\parsep}{0pt}
    \setlength{\topsep}{0pt}
    \setlength{\partopsep}{0pt}
    \setlength{\leftmargin}{2em}
    \setlength{\labelwidth}{1.5em}
    \setlength{\labelsep}{0.5em} } }

\newcommand{\squishend}{
  \end{list}  }

\newcommand{\ay}[1]{\textcolor{purple}{Amir: #1}}

\newcommand\circled[1]{\tikz[baseline=(char.base)]{
            \node[shape=circle,fill=black,inner sep=0.5pt] (char) {\textcolor{white}{#1}};}}

\usepackage{enumitem} 

\newenvironment{itempacked}{%
\begin{itemize}[noitemsep,nolistsep,leftmargin=*]
}
{%
\end{itemize}
}

\newcommand{\mdgf}{\textsc{MdGf}\xspace}
\newcommand{\sdgf}{\textsc{SdGf}\xspace}

\icmltitlerunning{Progressive Gradient Flow for Robust N:M Sparsity Training in Transformers}

\usepackage[capitalize,noabbrev]{cleveref}

\begin{document}

\twocolumn[
\icmltitle{Progressive Gradient Flow for Robust N:M Sparsity Training in Transformers}

% It is OKAY to include author information, even for blind
% submissions: the style file will automatically remove it for you
% unless you've provided the [accepted] option to the icml2024
% package.

% List of affiliations: The first argument should be a (short)
% identifier you will use later to specify author affiliations
% Academic affiliations should list Department, University, City, Region, Country
% Industry affiliations should list Company, City, Region, Country

% You can specify symbols, otherwise they are numbered in order.
% Ideally, you should not use this facility. Affiliations will be numbered
% in order of appearance and this is the preferred way.
\icmlsetsymbol{equal}{*}

\begin{icmlauthorlist}
\icmlauthor{Abhimanyu Rajeshkumar Bambhaniya}{equal,yyy}
\icmlauthor{Amir Yazdanbakhsh}{equal,comp}
\icmlauthor{Suvinay Subramanian}{sch}
\icmlauthor{Sheng-Chun Kao}{sch}
\icmlauthor{Shivani Agrawal}{sch}
\icmlauthor{Utku Evci}{comp}
\icmlauthor{Tushar Krishna}{yyy}
%\icmlauthor{}{sch}
% \icmlauthor{Firstname8 Lastname8}{sch}
% \icmlauthor{Firstname8 Lastname8}{yyy,comp}
%\icmlauthor{}{sch}
%\icmlauthor{}{sch}
\end{icmlauthorlist}

\icmlaffiliation{yyy}{Georgia Institute of Technology}
\icmlaffiliation{comp}{Google DeepMind}
\icmlaffiliation{sch}{Google}

\icmlcorrespondingauthor{Amir Yazdanbakhsh}{ayazdan@google.com}

% You may provide any keywords that you
% find helpful for describing your paper; these are used to populate
% the "keywords" metadata in the PDF but will not be shown in the document
\icmlkeywords{Machine Learning, ICML}

\vskip 0.3in
]

\printAffiliationsAndNotice{\icmlEqualContribution}

\begin{abstract}
% sparsity is interesting
% Sparsity has become one of the promising methods to compress and accelerate Deep Neural Networks (DNNs).
%
N:M Structured sparsity has garnered significant interest as a result of relatively modest overhead and improved efficiency.
%
% In particular, N:M sparsity is attractive because of hardware accelerator architectures capable of harnessing specific variations of N:M structured sparsity, enhancing computational efficiency.
%
Additionally, this form of sparsity holds considerable appeal for reducing the memory footprint owing to their modest representation overhead.
There have been efforts to develop training recipes for N:M structured sparsity, they primarily focus on low-sparsity regions ($\sim$50\%).
Nonetheless, performance of models trained using these approaches tends to decline when confronted with high-sparsity regions ($>$80\%).
In this work, we study the effectiveness of existing sparse training recipes at \textit{high-sparsity regions} and argue that these methods fail to sustain the model quality on par with low-sparsity regions.
We demonstrate that the significant factor contributing to this disparity is the presence of elevated levels of induced noise in the gradient magnitudes.
To mitigate this undesirable effect, we employ decay mechanisms to progressively restrict the flow of gradients towards pruned elements.
% present two new sparse training recipes, namely \emph{``Mask Decay Gradient Flow (\mdgf)''} and \emph{``Structure Decay Gradient Flow (\sdgf)''} which employ decay mechanisms to progressively restrict the flow of gradients.
%
% Our results demonstrate that enabling the propagation of gradients plays a crucial role in preserving superior model performance while simultaneously attaining a high level of sparsity.
%
% Our evaluations demonstrate that our methods consistently achieve SOTA accuracy against conventional recipes in a range of transformer models.
%
Our approach improves the model quality by up to 2$\%$ and 5$\%$ in vision and language models at high sparsity regime, respectively.
We also evaluate the trade-off between model accuracy and training compute cost in terms of FLOPs.
At iso-training FLOPs, our method yields better performance compared to conventional sparse training recipes, exhibiting an accuracy improvement of up to 2$\%$.
The source code is available at \href{https://github.com/abhibambhaniya/progressive_gradient_flow_nm_sparsity}{GitHub}.
% We have open-sourced our verified implementation and it  can be found at  \url{https://github.com/abhibambhaniya/progressive_gradient_flow_nm_sparsity}. 
%
\end{abstract}
\section{Introduction}
%
% Deep Neural Networks (DNNs) have achieved notable success in many domains, such as computer vision, language understanding, and machine translation.
% %
A prevailing tendency in state-of-the-art DNN models is the rapid increase in their model~\cite{t5_model,meta_opt,openai2023gpt4,touvron2023llama,team2023gemini}.
%
% For example, T5 from Google~\cite{t5_model}, OPT from Meta~\cite{meta_opt}, and GPT-4 from OpenAI~\cite{openai2023gpt4} have over 100 billion parameters.
%
% The exponential increase in model size poses significant obstacles to the deployment of these models, particularly in devices with limited computational resources.
%
To address the deployment challenges of these models, a large body of research proposes quantization~\cite{shen2020q, kim2021bert, zafrir2019q8bert, zhang2020ternarybert}, sparsification~\cite{evci2019difficulty,han2015deep, guo2016dynamic,he2017channel,molchanov2016pruning,yao2019balanced,zhu2017prune}, and distillation~\cite{Gou_2021}.
This paper centers its attention on \textit{sparsification/pruning} offering the following benefits: (a) improved performance~\cite{sr_ste}, (b) reduce memory usage~\cite{qin2021extending}, and (c) higher energy efficiency~\cite{akhlaghi2018snapea,pan2023bitset}.
%
% The benefits of sparsification are multi-fold.
%
% Firstly, sparsification decreases the computational requirements by avoiding multiplications involving pruned weights. 
% %
% Secondly, it reduces the memory usage by employing compressed sparse representations, enabling the deployment of large models in resource-limited devices~\cite{seshadri2022evaluation}.
% %
% Lastly, sparsification contributes to energy savings by eliminating unnecessary memory accesses for pruned weights and bypassing ineffectual computations.
%

While appealing, sparsification predominantly revolves around the inherent trade-offs between the quality of the model and compression ratio\footnote{We designate algorithmic-wise factors such as accuracy, recall, and precision as \textit{model quality.} and denote model runtime/latency as \textit{model performance}.}.
%
% \ay{Are they unstructured right? if yes fix.}
For example, some studies~\cite{guo2016dynamic, han2015learning} have demonstrated promising results in achieving unstructured sparsity levels of around 90$\%$-95$\%$ in image classification models, while maintaining the quality of dense models.
Similarly, the noticeable achievements of transformer-based models, primarily driven by their exponential growth in model size~\cite{wei2022emergent}, have stimulated interest~\cite{sparse_transformer,beltagy2020longformer,roy2020efficient,kitaev2020reformer} in exploring sparsification recipes for such models with high sparsity ratio.
This serves as a significant incentive for the sparsification of attention-based models, as it enables the pruning of a substantial number of model parameters (>70$\%$)~\cite{Efficient_Transformers,jaszczur2021sparse}.
% unstructured vs. structured
Despite its inherent ability to trim the memory footprint of large models, the realization of unstructured sparsity in hardware poses nontrivial challenges for acceleration.
%
% That is, the irregularity in the sparsity pattern hinders the efficient execution of sparse models by natively-dense accelerators such as GPUs and TPUs~\citep{tpu}.
%
The sparsity-induced models frequently exhibit comparable or inferior performance to their dense counterparts because of the additional intricacies involved in compression/decompression of model parameters~\cite{gpu_ampere,non_structured_pruning,renda2020comparing,lin2021filter,gamboa2020campfire,zhu2019sparse}.

As such, structured sparsity has gained significant popularity because of its hardware-friendly characteristics.
% , with a focus on regulating sparsity patterns such as channel/filter sparsity~\citep{li2016pruning,wen2016learning,he2017channel} or block sparsity~\citep{non_structured_pruning, pool2021channel,mishra2021accelerating,nvidia_asp, sr_ste}.
% %
% For example, natively-dense accelerators can simply bypass an entire channel computation without requiring any hardware modifications.
% %
% The caveat, however, is that structured sparsity generally entails a higher magnitude of quality loss.
%
\cite{yao2019balanced,kang2019accelerator, parashar2017scnn, s2ta, vegeta, bambhaniya2023accelerating, qin2022enabling} found that employing fine-grained N:M structured sparsity, has the potential to mitigate the degradation in quality.
Moreover, the debut of 2:4 structured-sparse tensor core in GPU Ampere architecture~\citep{gpu_ampere} has generated additional enthusiasm in developing efficient N:M training recipes.
Although recent methods~\citep{pool2021channel, mishra2021accelerating, nvidia_asp, sr_ste, lu2023step,frantar2023sparsegpt} demonstrate acceptable quality, their main focus lies in addressing sparsity levels up to 2:8.
These methods, however, less effective when dealing with high sparsity regimes such as 1:16, 1:32, and higher.
Through our studies, we identify that elevated levels of induced noise in the gradient magnitudes constitute a notable contributing factor to such quality degradation.
This phenomenon can be primarily attributed to either the absence~\citep{Stochastic_gradient_descent, Variance_Reduction} or perturbation of gradient flow of existing sparse training recipes.
Building on the insights our experiments, we introduce alternative training recipes that demonstrate substantial improvements in model quality, particularly at high sparsity regime.
We made the following contributions:
\begin{itempacked}
\item \niparagraph{The impact of gradient perturbations becomes increasingly evident at elevated levels of sparsity, leading to a deterioration in the quality of the model.} 
We present empirical evidence that SR-STE, a state-of-the-art N:M structured training recipe~\citep{sr_ste}, is less effective at high sparsity regions, $>75\%$.
We attribute this to the nontrivial perturbation of gradient magnitudes.
This perturbation during the initial stages of training\footnote{Recent studies for dense models~\citep{lu2023step, Stochastic_gradient_descent} have shown that the early stage of training (critical region) is imperative in the quality of training recipes.} adversely amplifies the variance of gradients, resulting in a diminished model quality.
\item \niparagraph{Gradient flow is all you need.}
In order to alleviate the adverse effects of noisy gradients, we introduce a class of decaying-based sparse training recipes tailored for N:M structured sparsity.
% : (1) \textit{Mask Decay Gradient Flow} (\mdgf) and (2) \textit{Structure Decay Gradient Flow} (\sdgf).
%
The fundamental principle underlying these methods involves progressively limiting the flow of gradients for \textit{pruned weights}, while allowing the gradients to freely flow at the early stages of training.
%
% \mdgf gradually decays the sparsity mask, either linearly or exponentially, instead of employing the conventional binary mask.
% %
% In contrast, \sdgf encompasses a collection of iterative pruning methods that progressively increases the sparsity ratio, while maintaining the overall N:M sparsity patterns.
%
% We compare the proposed methods against SR-STE across range of transformer models and under different sparsity configurations.
%
Our results demonstrate that the decaying-based methods consistently outperform SR-STE by up to 2$\%$-5$\%$ in terms of model quality, while pruning $\sim$97$\%$ of parameters.
\item \niparagraph{Decaying-based sparse training recipes require less training FLOPs.}
To better understand the computational overhead of the proposed sparse training recipes, we present the trade-off between model accuracy and training compute cost in term of FLOPS.
The results show that at iso-quality, our method requires $>30\%$ fewer training FLOPs compared to SR-STE.
% iso-training FLOPS, the proposed method delivers up to 2.0$\%$ better model quality. 
% %
% That is, at iso-quality, \sdgf requires 
%
\end{itempacked}
% \begin{figure*}
% \begin{center}
% \includegraphics[width=0.8\linewidth]{Figures/dense_sparse.pdf}
% \end{center}

% \caption{The computation flow of (a) Dense training, (b) Sparsify, (c) Fine-tuning, and (d) Sparse training. The training schedule of (e) regular dense training, (f) fine-tuning with one-shot sparsifying, (g) fine-tuning with iterative sparsifying, (h) from-scratch with learned one-shot sparsity pattern, and (i) from-scratch while learning sparsity pattern. The sparsify algorithm in (d):  Evolutionary Strategy(ES) such as \citep{mocanu2018scalable}, Bayesian Optimization such as \citep{bellec2017deep}, Trainable parameters such as \citep{dettmers2019sparse, kusupati2020soft}.}

% \label{fig:dense_sparse}
% \end{figure*}

\section{Background and Related Works}
This work  focuses on weight sparsity, which poses a significant challenge in serving transformer-based models.
\subsection{Computation Flow of Sparse Training Recipes}
\autoref{fig:dense_sparse} summarizes the computation flows of various training recipes for the sparsification of model parameters.
A sparsification recipe broadly entails 1) pruning criteria, 2) pruning schedule, and 3) sparsity pattern.

\niparagraph{(1) Pruning criteria.}
The pruning criteria refers to the set of criteria used to determine the specific elements within the weight tensor that should be pruned.
Magnitude pruning, which selects the pruning elements based on their absolute values, is one of the most widely used criteria for sparsification~\cite{renda2020comparing,guo2016dynamic,lee2018snip,frankle2018lottery,gale2019state,zhu2017prune,han2015deep,liu2018rethinking}.
Recent work employs other criteria such as gradient~\cite{yeom2021pruning,evci2020rigging}, Hessian~\cite{lecun1989optimal}, connection sensitivity~\cite{lee2018snip}, and importance estimation~\cite{molchanov2019importance}.
In this paper, we use magnitude pruning, following SR-STE~\cite{sr_ste} the state-of-the-art structured N:M training recipe.

\niparagraph{(2) Pruning schedule.}
We classify the pruning schedules into the following broad categories:
\begin{itempacked}
\item \hl{Fine-tuning with one-shot pruning}$\rightarrow$ This approach~\cite{mishra2021accelerating,pool2021channel, frankle2018lottery,lee2018snip} involves training a dense model, followed by on-shot weight pruning. Subsequently the pruned model is fine-tuned to regain the lost quality.
\item \hl{Fine-tuning with iterative pruning}$\rightarrow$ This method~\cite{evci2019difficulty,han2015deep, guo2016dynamic,he2017channel,molchanov2016pruning,yao2019balanced,zhu2017prune,gamboa2020campfire,narang2017exploring,narang2017block,elsen2020fast,evci2020rigging} trains a dense model followed by iterative cycles of pruning and re-training, which shows a greater capacity to regain lost quality.
\item \hl{From-scratch with learned pruning pattern}$\rightarrow$~This pruning recipe~\cite{frankle2020pruning,evci2019difficulty} establishes the sparsity pattern based on pretrained dense model and subsequently trains a sparse model from scratch.
% , as shown in \autoref{fig:dense_sparse}h.
% %
% % \ay{are you sure evci doing this? is this like dense pretraining and sparse fine-tuning? ->                      AB:Yes Fig2 of the paper shows that}
% %
% This pruning schedule establishes the sparsity pattern based on the pretrained dense model and subsequently trains a sparse model from scratch. 
%
\begin{figure*}
\begin{center}
\includegraphics[width=0.99\linewidth]{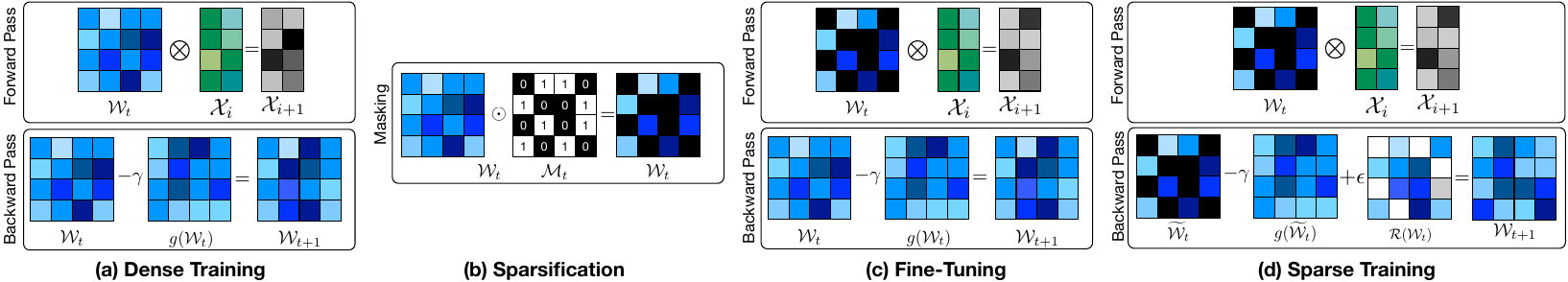}
\end{center}
\caption{The computation flow of (a) dense training, (b) sparsification, (c) fine-tuning, and (d) sparse training (e.g. SR-STE). $\mathcal{\widetilde{W}}$ represents a pruned matrix that is computed by element-wise multiplication ($\odot$) of $\mathcal{W}$ and its sparsification mask ($\mathcal{M}$).
Sparse training recipes, such as SR-STE, introduce a ``\textit{sparse refining}'' regularizer ($\mathcal{R}$) to adjust the gradient terms for pruned elements.}
\label{fig:dense_sparse}
\end{figure*}
\item \hl{From-scratch while learning sparsity pattern}$\rightarrow$ 
This approach~\cite{wortsman2019discovering,dettmers2019sparse,gale2019state,kusupati2020soft,evci2020rigging,bellec2017deep,mocanu2018scalable} trains a sparse model from scratch while concurrently learning the sparsity mask.
% , as shown in \autoref{fig:dense_sparse}i.
% %
% This method trains a sparse model from scratch while concurrently learning the sparsity pattern.
\end{itempacked}
% \textit{i) Fine-tuning with one-shot pruning} (\autoref{fig:dense_sparse}f)~\cite{mishra2021accelerating,pool2021channel, frankle2018lottery,lee2018snip}, which trains a dense model, prunes the weight with one-shot and re-trains the model to recover the quality loss.
%
% \textit{ii) Fine-tuning with iterative pruning} (), which trains a dense model and then iterates between pruning and re-training. These schemes usually have a higher ability to recover the quality loss.
%
% \textit{iii) From-scratch with learned one-shot pruning pattern} (\autoref{fig:dense_sparse}h)~\cite{frankle2020pruning,evci2019difficulty}, 
% \textit{iv) From-scratch while learning sparsity pattern} (\autoref{fig:dense_sparse}i), 

\niparagraph{(3) Sparsity pattern.}
We broadly categorize sparsity patterns into following groups:

\begin{itempacked}
    \item \hl{Unstructured Sparsity} refers to the process of pruning a model without imposing  any constraints on the sparsity pattern~\cite{renda2020comparing,guo2016dynamic,lee2018snip,frankle2018lottery,gale2019state}.
    This sparsity pattern is known to be able to prune the model size to an order of magnitude smaller while retaining a similar model quality as its dense counterpart at the cost of increased runtime overhead.
    \item \hl{Coarse-grained Structured Sparsity} enforces coarse-grained sparsity patterns, including techniques like filter/channel pruning~\cite{li2016pruning,wen2016learning,he2017channel} and block-wise pruning~\cite{wen2016learning,non_structured_pruning,narang2017block,gray2017gpu}.
    By skipping the entire computation of a tensor, this sparsity pattern often yields speedup in natively-dense accelerators such as GPUs and TPUs.
    Nevertheless, this trade-off often results in a reduction in model quality.
    \item \hl{Fine-grained Structured N:M Sparsity} prunes (M-N) out of M consecutive elements.
    Several preliminary studies rely on special threading and grouping techniques~\cite{yao2019balanced} or specialized sparse accelerators~\cite{kang2019accelerator} to exploit this fine-grained sparsity pattern.
    With the inclusion of 2:4 GEMM support in GPU Ampere architecture~\cite{gpu_ampere}, recent work starts to investigate effective training recipes for N:M sparsity patterns to harness the existing accelerators~\cite{pool2021channel,mishra2021accelerating,nvidia_asp,sr_ste}.
\end{itempacked}

% \subsection{Sparsification of Attention Models} 
% %
% \ay{The organization here was not great.}

% \niparagraph{Profiling attention models.}
% %
% To better understand the characteristics of attention models for sparsification, we profile the ViT-Base model on TPUv3.
% %
% \ay{I don't get the following sentence!Einsum!}
% %
% We observe the three major components of the ViT-Base model are:
% %
% \textcircled{1} ``Einsum'' $\rightarrow$ computation of attention scores and the weighted sum of values by the attention scores.
% %
% \textcircled{2} ``Projections'' $\rightarrow$ projecting inputs to key(K), query(Q), and value(V), and projection weighted sum of values \ay{values should be attention scores?} to outputs(O). 
% %
% \textcircled{3} ``Feed Forward (FFs)'' $\rightarrow$
% %
% The feed-forward layers at the end of the attention layer. 
% %
% Other layers and operations such as ReLU, LayerNorm, Add, Softmax, embedding have little contributions to the FLOPs and size of model parameters, hence not included in the estimation. 
% %
% This limited study demonstrates that the feed-forward layers account for around 64\% of overall FLOPs and 63.6\% of parameters\footnote{As the sequence length increases, the impact of FF layers on FLOPs and parameter count becomes more pronounced\cite{kim2023stack}.}.
% %
% Therefore, we primarily focus on studying the sparsification of FF layers in transformer-based models.
%
\begin{figure*}
\begin{center}
\includegraphics[width=0.99\linewidth]{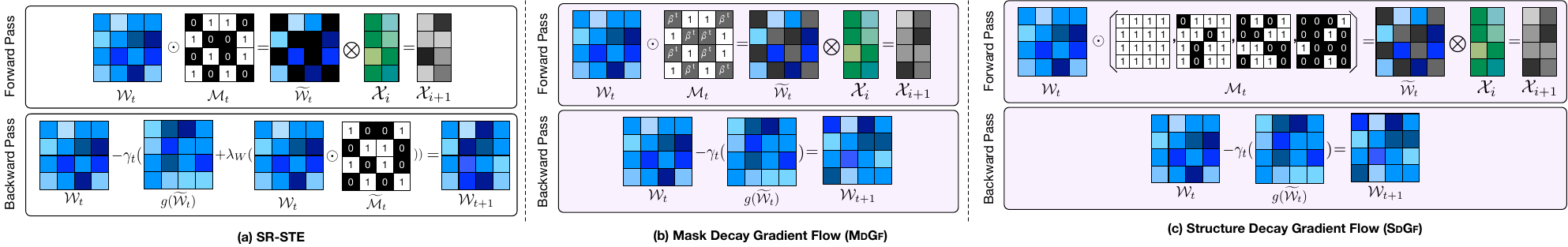}
\end{center}
\caption{An overview of different sparse training recipes (a) SR-STE~\cite{sr_ste}, (b, c) proposed decaying mechanisms in this work. (b) indicates decaying binary mask values for pruned weights (\mdgf), whereas (c) gradually change the N:M sparsity patters at different intervals (\sdgf).}
\label{fig:alg_flow}
\end{figure*}

\niparagraph{Other related work.}
Other work has also investigated N:M structured sparsity in attention-based models.
SR-STE~\cite{sr_ste} proposes a training recipe with fine-grained N:M structured sparsity from scratch.
\autoref{fig:alg_flow}(a) demonstrates the weight update scheme for the forward and backward pass of SR-STE.
Nvidia ASP~\cite{nvidia_asp} focuses on low sparsity (2:4) and employs channel permutations to maximize the accuracy of N:M sparse networks.
However, this approach becomes slow for higher sparsification levels because of the lack of hardware support.
SparseGPT~\cite{frantar2023sparsegpt} introduces a post-training sparsification recipe tailored for GPT-family models. 
SparseGPT shows on-par model quality with up to 50$\%$ weight pruning under unstructured and N:M structured sparsity.
%
% Orthogonal to weight sparsification methods, \cite{Efficient_Transformers} centers around activation sparsity in order to reduce model complexity.
% %
% \ay{This sentence needs some rewriting. ``which penalizes weights targeted for pruning.''???}
%
Finally, selective weight decay (SWD)~\cite{Tessier_2022} is a pruning method based on Lagrangian smoothing, which penalizes weights that are selected for pruning.
However, SWD neither explores attention models nor provides training recipes for N:M structured sparsity.
\section{Decaying-based Sparse Training Recipes}
This section covers the class of decaying-based training recipes for fine-grained N:M sparsity.
The main premise of these recipes is to allow the gradient to flow through weight tensors in a controlled way to prevent induced noise in the gradients.
We broadly classify the proposed decaying-based training recipes into: (a) ``\underline{\textbf{M}}ask \underline{\textbf{D}}ecay \underline{\textbf{G}}radient \underline{\textbf{F}}low'' (\mdgf) and (b) ``\underline{\textbf{S}}tructure \underline{\textbf{D}}ecay \underline{\textbf{G}}radient \underline{\textbf{F}}low'' (\sdgf), each with sub-variants which we discuss in details below.
In contrast to~\citep{sr_ste}, we intentionally refrain from modifying the gradient update rules in either of these categories.
Instead, we use different update rules for sparsity pattern or sparsity mask tensor, facilitating unimpeded gradient flow during the entire sparse training process.

\niparagraph{Implementation.}
In order to implement these methods, we employ the process of pruning dense weight tensors ($\mathcal{W}_t$) to generate sparse weight tensors ($\widetilde{\mathcal{W}_t}$), adhering to the following rule during the forward pass:
\begin{align*}
    \widetilde{\mathcal{W}} &= \mathcal{F}(\mathcal{W}, N, M, \Phi, \beta, j) \\
    &= \mathcal{W} \odot [\Phi(\mathcal{W}, N, M, j) + \mathcal{D}(j)(1-\Phi(\mathcal{W}, N, M, j))]
\end{align*}
Here $\odot$ represents the Hadamard product.
$\Phi(\cdot)$ and $\mathcal{D}(\cdot)$ calculate a decaying-based binary mask and decay mask factor, respectively.
Each function's implementations establish distinct decaying-based training recipes.
$\Phi(\cdot)$ calculates a binary mask that matches the dimensions of the input weight tensor ($\mathcal{W}$). 
The location of 0s and 1s in the binary mask refers to pruned and unpruned weights, respectively.
In fine-grained N:M structured sparsity with magnitude pruning, $\Phi(\cdot)$ assigns a value of 1 to the N weight tensor elements with the highest absolute magnitude within a contiguous block of M elements.
Simultaneously, it enforces all the other elements with the block to be set to 0.
%
% In all of our experimental setups, we induce N:M sparsity along the row dimension of the weight tensor.
% %
% Our results (supplementary material Section 1) shows that row-wise sparsity slightly outperforms column-wise sparsity.
%
In addition, $\mathcal{D}(\cdot)$ calculates the decaying factor for binary mask according to the target decaying-based training recipe.

\niparagraph{\circled{1} \underline{M}ask \underline{D}ecay \underline{G}radient \underline{F}low (\mdgf).}
In the first training recipe~\autoref{fig:alg_flow}~(b), we propose the use of a diminishing value ranging from 1 to 0, as opposed to the commonly-used binary pruning mask (e.g., ``0'' $\rightarrow$ pruned and ``1'' $\rightarrow$ dense).
Note that for the mask-decay training recipes the function $\Phi(\cdot)$ produces a mask tensor either with all ones (dense training) or with a sparsity pattern following target N:M fine-grained structured sparsity. 
In the initial epochs, we use a mask comprising solely of ones and assign a constant value of 1 to $\mathcal{D}(\cdot)$, i.e., dense training.

Upon staring sparse training phase, $\mathcal{D}(\cdot)$ produces gradually diminishing floating-point values between 1 and 0.
The output of function $\mathcal{D}(\cdot)$ depends on current decaying interval. 
Using a diminishing decaying factor enables gradient flow for both pruned and unpruned weights.
This is in contrast to prior work in which $\mathcal{D}(\cdot)$ is null which may cause instability in the training process.
We propose two alternative implementations for $\mathcal{D}(\cdot)$ as follows:
\begin{itempacked}
\item \mdgf-Linear uses $\mathcal{D}(j) = max(1 - K_{\tau}\times j, 0)$ that reduces the decay mask values linearly with respect to training steps.
\item \mdgf-Exponential, as its name implies, we use $\mathcal{D}(j) = e^{-K_{\eta}\times{j}}$, indicating an exponential decrease in the mask decay value relative to the ongoing training step.
\end{itempacked}
The value of $K_{\tau/\eta}$ determines the rate of decay.
To ensure a binary mask value for the target N:M sparsity pattern, after sufficient decaying intervals, $\mathcal{D}(\cdot)$ approaches zero.
After reaching the target N:M sparsity pattern, we proceed with few additional training epochs to restore the model accuracy.
We postulate that using non-binary pruning mask values facilitates the smooth propagation of gradients in pruned weights, resulting in more stable sparse training.
%
% \begin{figure*}
%   \centering
%   \begin{minipage}{0.4\textwidth}\centering
%     \includegraphics[width=\textwidth]{Figures/MDGF-linear.png}
%     \subcaption{(a) \mdgf-Linear}
%   \end{minipage}
%   % \hfill
%   \begin{minipage}{0.4\textwidth}\centering
%     \includegraphics[width=\textwidth]{Figures/MDGF-exp.png}
%     \subcaption{(b) \mdgf - Exponential}
%   \end{minipage}

%   \vfill

%   \begin{minipage}{0.4\textwidth}\centering
%     \includegraphics[width=\textwidth]{Figures/SDGF-Stepwise.png}
%     \subcaption{(c) \sdgf-Stepwise}
%   \end{minipage}
%   % \hfill
%   \begin{minipage}{0.4\textwidth}\centering
%     \includegraphics[width=\textwidth]{Figures/SDGF-Geometric.png}
%     \subcaption{(d) \sdgf-Geometric}
%   \end{minipage}
% %
% \caption{The sparsification schedules for \mdgf and \sdgf variants. The variables $\beta^t$ and N:M  are same as explained in \autoref{fig:alg_flow}(b). \ay{update the figure to be exactly the same as the configurations we used, whether it is 8 or 16.}\AB{doing right now}} 
%
% \label{fig:structure_schedule}
% \end{figure*}

%

\niparagraph{\circled{2} \underline{S}tructure \underline{D}ecay \underline{G}radient \underline{F}low (\sdgf).}
\sdgf decays the structure of the pruning mask, e.g. gradually altering the sparsity level, e.g. $\frac{3}{4}\mapsto\cdot\cdot\cdot\mapsto\frac{1}{4}$.
In contrast to \mdgf, this method strictly confines the pruning mask values to either 1 or 0, e.g. $\mathcal{D}(\cdot) = 0$.
We propose two alternative implementations of $\Phi(\cdot)$, (a) \textit{Stepwise} and (b) \textit{Geometric}.

The \textit{\sdgf-Stepwise} starts by inducing M-1:M structured sparsity.
Subsequently, it gradually increase the level of sparsity following $\frac{M}{2^d}:M$ formulation in which $d$ denotes the index of the decaying interval, until $\frac{M}{2^d} == N$.
For example, to retain a target sparsity level of 1:8, the method applies the following sparsity patterns at different decaying interval $ \frac{7}{8} \mapsto \frac{4}{8} \mapsto \frac{2}{8} \mapsto \frac{1}{8}$. 

The core idea of \textit{\sdgf-Geometric} is to maintain a constant ratio of $\frac{N}{M}$ throughout successive decay intervals by adjusting the values of N and M in proportion to each other.
In all experiments, we configure $\Phi(\cdot)$ to be $\frac{k\times{M}}{2^d}:\frac{k\times{N}}{2^d}$.
The value of $k$ is set to 16, unless specifies otherwise.
We empirically find that $k > 16$ offers negligible improvements in terms of model quality.
For example, for a target sparsity of 1:8, we induce the following sparsity patterns at each decaying interval, $ \frac{16}{128} \mapsto \frac{8}{64} \mapsto \frac{4}{32} \mapsto \frac{2}{16} \mapsto \frac{1}{8}$.
For both recipes, we evenly partition the total sparsification epochs throughout the decaying intervals.
Fundamentally, this approach follows a hypothesis akin to \mdgf.
Enabling the flow of gradients of pruned weights throughout the model potentially leads to higher model accuracy. 
\begin{figure*}[t]
   \centering
\subfigure[Variance of AdamW Second Moment]{\label{fig:gradient_std:a}
\includegraphics[width=0.35\linewidth]{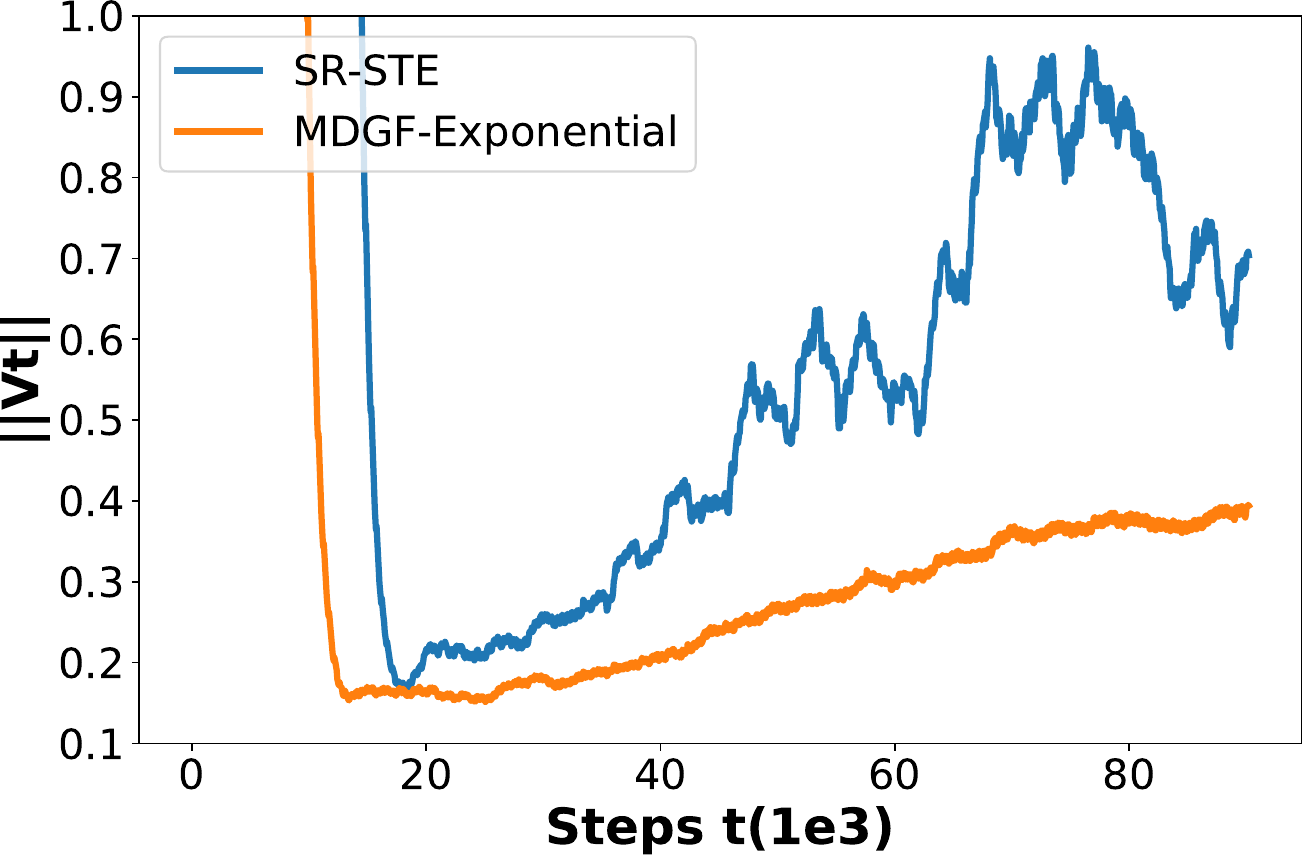}}
% \hfill
\hspace{1cm}
\subfigure[Gradient Variance]{\label{fig:gradient_std:b}
\includegraphics[width=0.35\linewidth]{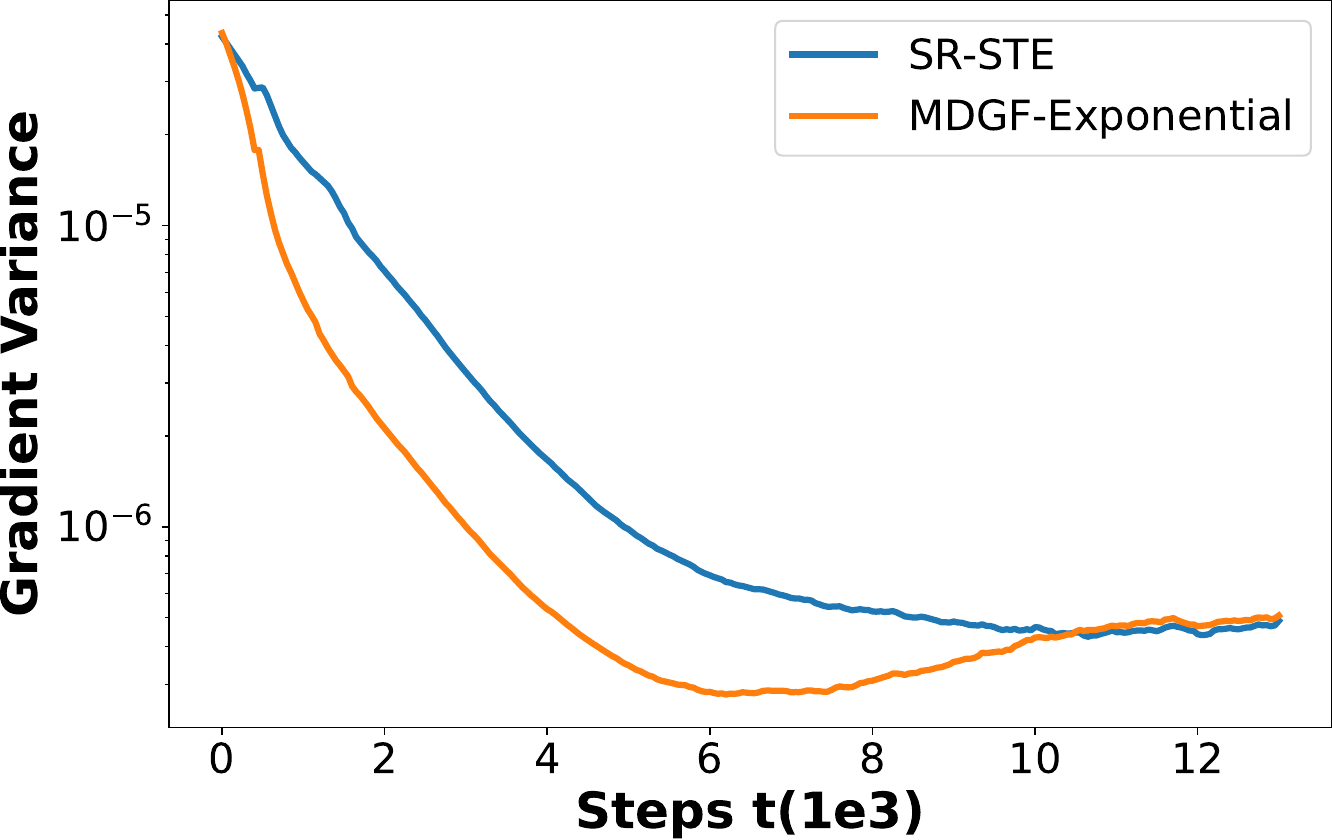}}
\caption{Trends for different indicators of gradient values during training. Data from ViT-tiny trained on CIFAR-10 with 1:16 sparsity pattern. (a) and (b) show the running average of the variance of AdamW second moment and gradient variance, respectively.}

\label{fig:gradient_std}
\end{figure*}

\section{Impact of Gradient Flow in Sparsification}
To gain better understanding of the advantages of proposed decay methods, we conducted an empirical analysis to compare the gradient values of \mdgf-Exponential and SR-STE~\cite{sr_ste}.
We created a compact version of ViT with three encoder layers, each with three attention heads, and an embedding size of 192.
We trained this model on CIFAR-10~\citep{cifar10} for 200 epochs with batch size 64 with AdamW optimizer.
To understand the impact of sparsification, we collect and analyze two different metrics, namely \textit{second moment} and \textit{gradient variance}. These values are an indication of how effective the gradient estimations are for training~\cite{tang20211,lu2022maximizing,li20221}. 

\subsection{Analysis of Second Moment Estimates}
\autoref{fig:gradient_std:a} shows the variance of the second moment term (exponential moving average of squared gradient values) for Feed-Forward (FF) layers in the model.
We observe that in \mdgf, the variance steadily decreases in magnitude, whereas in SR-STE, the variance stays at the relatively high level even at the later stages of training.
Prior study~\cite{tang20211,lu2022maximizing,li20221} correlate lower variance of the second moment with faster convergence rate during training and better model accuracy. This suggests that the gradient noise induced by SR-STE have negative impact on the convergence of the model and model accuracy.
\subsection{Analysis of Gradient Noise}
\autoref{fig:gradient_std}(b) shows the  variance of absolute back-propagation gradients. These values can be interpreted as the amount of noise in the gradient estimates. Similar to the previous study, we collect the gradients of Feed-Forward(FF) layer in tiny-ViT. 
We observe that in \mdgf, the variance of gradients decreases quickly, whereas in SR-STE, the variance of gradients has a lower slope (e.g. taking a larger number of steps).
When the variance of the gradient is higher, the optimizer spends time bouncing around, leading to slower convergence and lower performance~\cite{Stochastic_gradient_descent,Variance_Reduction}.
The variance for \mdgf-exponential comes down rather quickly thus the gradients are less noisy compared to SR-STE. This would result in higher accuracy for \mdgf-Exponential.
When observing the final validation accuracy of the two runs, we confirm our intuitive conclusions as the SR-STE accuracy is lower compared to \mdgf-Exponential accuracy.
%
% For \mdgf, the variance gradually becomes small in magnitude, in contrast, for SR-STE, the variance remains relatively large even in the later stages of training.
% %
% During the training process, the variance steadily decreases in magnitude for \mdgf, while in the case of SR-STE, the variance persists at a relatively high level even during the later stages of training.
% %
% \niparagraph{Conclusion.}
% %
% Lower variance of the second moment suggests the model converges faster and thus can achieve better accuracy when trained for a same number of epochs as a model that has not converged. %\citep{lu2023step,kingma2017adam} 
% %
% This suggests that noise gradients for SR-STE even during the later stages of training, would result in lower accuracy compared to the \mdgf-exponential.
\section{Experiment}
\label{sec:eval}
\begin{table*}[!htbp]\centering
\scriptsize
\caption{The compute and memory contributions of the three major layers in Transformers. These estimations are made for ViT-Base. The FF layers account for around 64\% of overall FLOPs and 66.6\% of parameters. We use sequence length 196 to read image of 224x224.}
\vspace{0.5em}
{\begin{tabular}{lccc}\toprule
&Einsum (Logit \& Attend) & Projections (Q/K/V/O) & Feed Forward (FF1/FF2)  \\\midrule
(T)FLOPS & 1.42~(4\%) & 11.1~(32\%) & 22.20~(64\%) \\
Params (MB) & 0.0~(0\%) & 28.31~(33.3\%) & 56.62~(66.6\%)\\
\bottomrule
\end{tabular}}
\label{table:compute_contribution}
\end{table*}

In this section, we evaluate the effectiveness of various training recipes for N:M fine-grained structured sparsity in a range of attention-based models and tasks, such as image classification, language translation and understanding.
\begin{table*}[!htp]\centering
\caption{ImageNet-1K Top-1 validation accuracy on \xx{ViT-Base} across different N:M sparsity patterns and training recipes.}
\vspace{0.5em}
\scriptsize
\resizebox{0.98\textwidth}{!}{
\begin{tabular}{l|c|ccccc}\toprule
Sparse Target & Dense & SR-STE	& \mdgf-Linear & \mdgf-Exponential	& \sdgf-Stepwise	& \sdgf-Geometric \\\midrule
% \multirow{6}{*}{\makecell{Sparsity\\ Target}}
2:4 (FF)	& 76.389 & \textbf{77.761} & 77.613 & 76.381 & 77.081 & 77.363 \\
1:4 (FF)	& 76.389 & \textbf{78.782} & 78.512 & 78.579 & 77.357 & 78.347 \\
1:8 (FF)	& 76.389 & 77.869 & 78.019 & 78.009 & 77.025 & \textbf{78.175} \\
1:16 (FF)	& 76.389 & 75.637 & 76.594	  & \textbf{77.325} & 75.923 & 76.869 \\
1:32  (FF)	& 76.389 & 73.056 & 75.807 & \textbf{76.068} & 74.394 & 74.910 \\
1:128 (FF)	& 76.389 & 72.069 & 74.012 & \textbf{74.180} & 71.725 & 69.801 \\
1:4   (FF) + 1:4 (QK) & 76.389 & 78.145 & 77.755 & 78.113 & 77.163 & \textbf{78.229} \\
1:8   (FF) + 1:8 (QK) & 76.389 & 75.527	& 76.473 & \textbf{77.349} & 76.617 & 76.334 \\
1:8   (FF) + 1:4 (QK) & 76.389 & 78.144	& 78.025 & \textbf{78.273} & 77.163 & 76.839 \\
1:8   (FF) + 1:4 (QKV)& 76.389 & 78.222	& 78.319 & \textbf{78.319} & 77.309 & 78.213 \\
\bottomrule
\end{tabular}
}
\label{table:vit_ff}
\end{table*}
Motivated by the relatively substantial contribution of FF layers (\autoref{table:compute_contribution}) in total FLOPs ($\sim$64$\%$) and parameter count ($\sim$66.6$\%$), we center our experiments around sparsification of these layers within the encoder and decoder blocks.
In addition, we conduct experiments on the pruning of projection layers ($\mathcal{Q}/\mathcal{K}/\mathcal{V}$) for a variant of \xx{ViT-Base}~\citep{vit}, a variant of \xx{SwinV2-Base}~\citep{liu2022swin}, and \xx{T5X-Base}~\citep{t5_model}.
For \xx{ViT-Base}, we use fixed-size patches (resolution 16$\times$16) on images with resolution 224.
In \xx{SwinV2-Base}, we employ window sizes of 8$\times$8 on images with  resolution 256.
For image classification tasks, we branched (commit: \href{https://github.com/huggingface/pytorch-image-models/commit/130458988a61c961cd78eb95c427472af5a26e50}{1304589}) our implementation from PyTorch Image Models~\citep{rw2019timm} and use NVIDIA A100 GPUs for training on ImageNet-1K dataset~\citep{imagenet}.
For \xx{T5X-Base}, we extend the official Google T5X release (commit: \href{https://github.com/google-research/t5x/commit/d3d3cbfc5c204244393f625b11d56f64f1138dbd}{d3d3cbf}) with sparsification training recipes and use Google TPUv3.
We train these models from scratch using different training recipes across different patterns of N:M fine-grained structured sparsity.   
SR-STE~\citep{sr_ste} serves as the baseline sparse training recipe  to assess the effectiveness of the proposed training recipes in terms of model accuracy.
\autoref{sec:sup_hyperparams_appx} have details about training hyperparameters, dataset details, and evaluation metrics.

\subsection{Image Classification \texorpdfstring{$\mapsto$ \xx{ViT-Base} and \xx{SwinV2}}{}}
%
% \subsection{Model Quality}
%
\begin{figure}
    \centering
    \includegraphics[width=\linewidth]{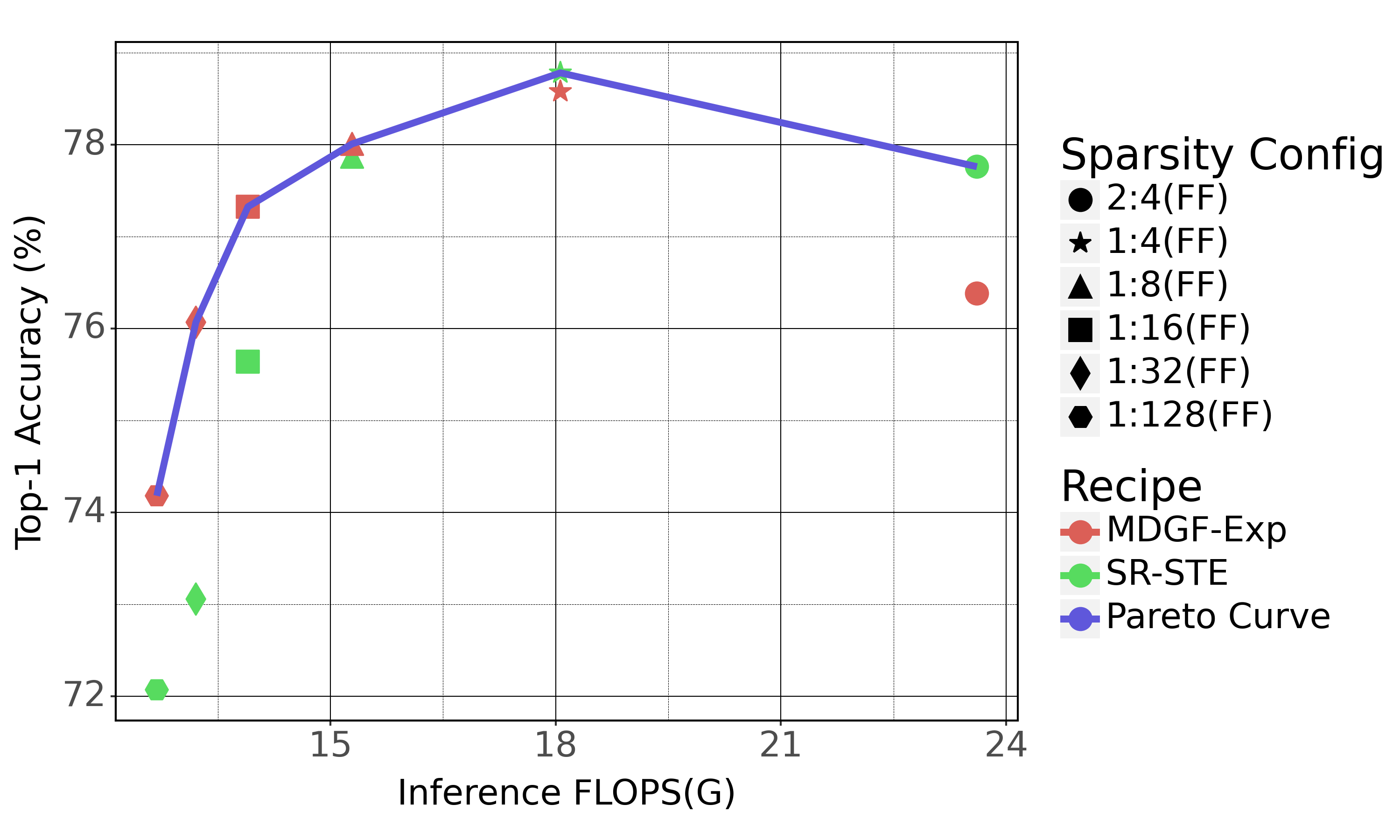}
    \caption{FLOP vs. Accuracy for ViT-Base+ImageNet-1K.}
    \label{fig:mac_accuracy}
\end{figure}
\niparagraph{\xx{ViT-Base} model quality.}
\autoref{table:vit_ff} presents Top-1 validation accuracy for variations of N:M sparsity in \xx{ViT-Base}, with the highest accuracy model indicated in bold.
The ``\textit{Sparse Target}'' column signifies the intended level of N:M sparsity.
For example, a sparsity target of 1:32 indicates that sparse tensors exhibit at most one non-zero for every 32 contiguous elements.
In low sparsity scenarios (e.g., 2:4 and 1:4), both \mdgf and SR-STE demonstrate comparable performance.
Nevertheless, with increases in either sparsity degree (e.g., 1:8 and higher) or the number of sparse layers, e.g., 1:4 ($\mathcal{FF}$) + 1:4 ($\mathcal{QK}$), employing SR-STE is detrimental to model quality.
In contrast, the proposed decaying-based training recipes, \mdgf and \sdgf, yield the highest accuracy.

\begin{figure*}[ht]
\centering
\subfigure[Accuracy vs. Sparsity ratio showing Occam’s hill.]
{\label{fig:occum_hill:a}
\includegraphics[width=0.48\linewidth]{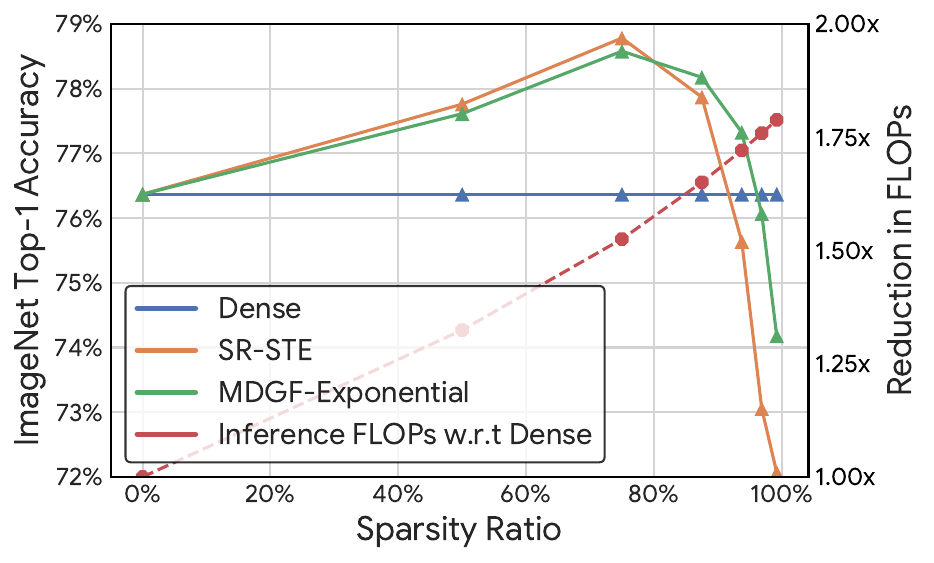}}
\subfigure[Accuracy vs \% of training epochs.]{\label{fig:occum_hill:b}
\includegraphics[width=0.42\linewidth]{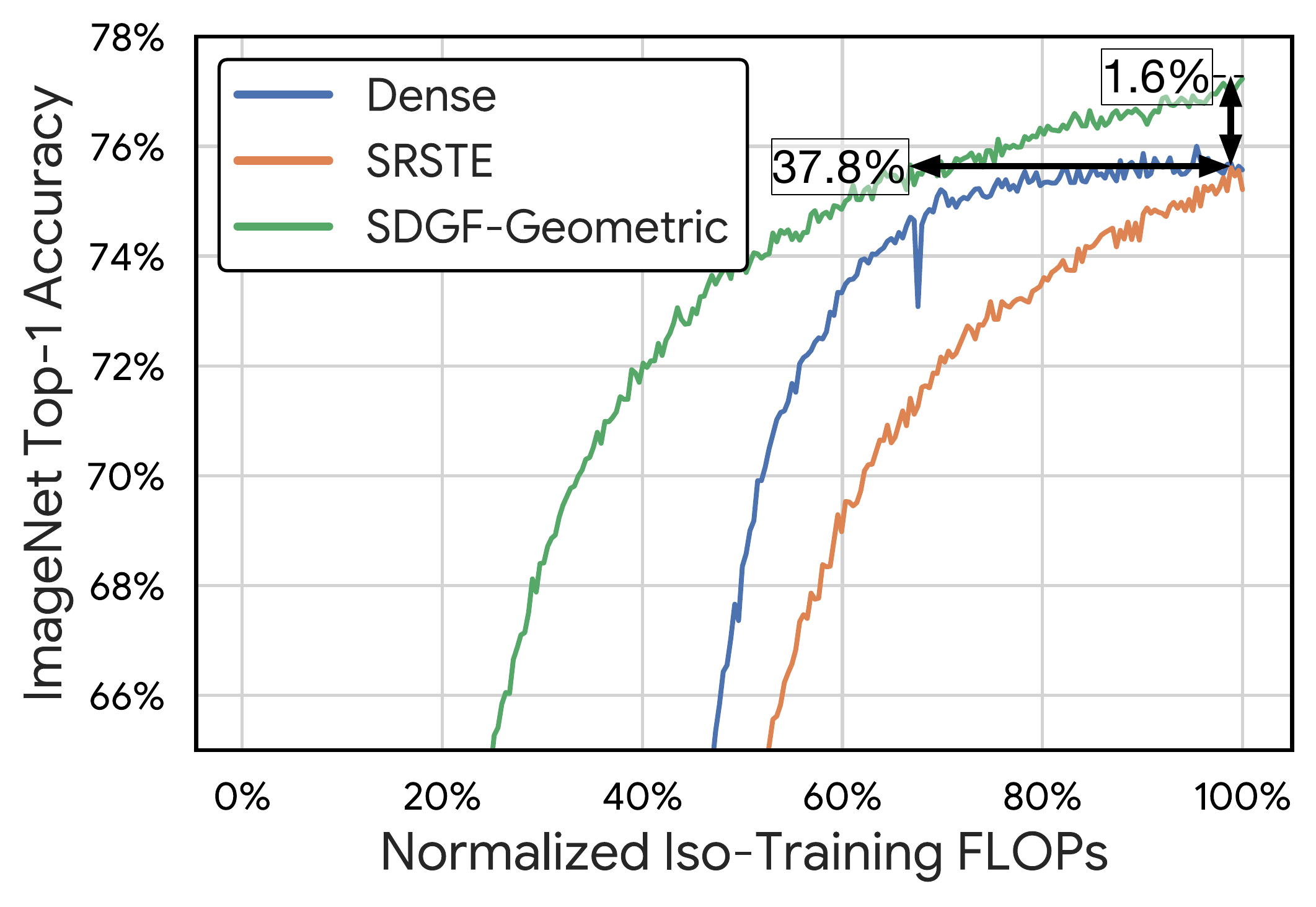}}
\caption{\xx{ViT-Base} trained on ImageNet-1K with different sparsity patterns and targets. (a) shows the Occam's hill where sparsity improves the model accuracy. The dashed red line shows the reduction in inference FLOPs at different sparsity ration. At high sparsity regime ($>$80\%) \mdgf yields better accuracy than SR-STE and (b) demonstrates model accuracy across training recipes (dense and sparse) at different training FLOPs. The vertical line indicates the proposed decaying method is better (1.6\%) than dense model at given training FLOPS. The vertical line shows that the decaying based method reaches to dense model accuracy at 37.8\% less training FLOPs.}
\label{fig:occum_hill_tot}
\end{figure*}
Interestingly, when aiming for a sparsity target of 1:32 (approximately 97$\%$), \mdgf-Exponential showcases a mere 0.3$\%$ reduction in accuracy compared to a fully dense model (76.389 vs. 76.068).
Additionally, we notice that the model accuracy increases at modest sparsity degrees, specifically in 2:4/1:4/1:8 (FF) patterns, resulting in an improvement of up to $\Delta(Acc)=+2.4\%$ in 1:4 (FF).
The increase in model accuracy, demonstrated in \autoref{fig:occum_hill:a}, can be attributed to Occam's Hill, wherein the positive impact of sparsity as a means of regularization is elucidated~\citep{Occams, hoefler2021sparsity}.
The performance of \mdgf-Exponential training recipe is comparable to that of SR-STE in low-sparsity scenarios.
However, the proposed \mdgf-Exponential recipe far surpasses SR-STE when confronted with high-sparsity patterns.

As commercially available accelerator can not support high-sparsity patterns.
In order to assess the \textit{potential} performance benefits by comparing the savings in inference FLOPs as well as memory usage.
\autoref{fig:mac_accuracy} visualizes the trade-off between accuracy and inference FLOPs across range of sparsity configurations and recipes.
The results show that \mdgf-Exponential with sparsity 1:16 provides similar accuracy as SR-STE 2:4 with 60\% fewer inference FLOPs and 30\% fewer parameters.
\autoref{sec:flops_calc} provides the details of FLOPs calculations.
% For details of FLOPs calculations, please refer to \autoref{sec:flops_calc}.

\niparagraph{\xx{SwinV2-Base} model quality.}
\autoref{table:swin_ff} demonstrate Top-1 validation accuracy for \xx{SwinV2-Base}.
Similar to \xx{ViT-Base}, we observe that the decaying-based algorithms outperforms SR-STE across various N:M sparsity patterns.
In 1:4 and 1:8~($\mathcal{FF}$), \sdgf-Geometric yields the highest Top-1 validation accuracy.
Whereas, in high-sparsity patterns, \mdgf-Exponential demonstrates superior performance compared to SR-STE.
To summarize, the results from the two image classification models demonstrate that the proposed training recipes, \mdgf and \sdgf, which incorporate decaying-based approaches for N:M fine-grained structured sparsity, yield superior performance compared to SR-STE.
\begin{table*}[!htp]\centering
\caption{ImageNet-1K Top-1 validation accuracy on \xx{SwinV2-Base} across different N:M sparse patterns and training recipes.}
\vspace{0.5em}
\scriptsize
\resizebox{0.8\textwidth}{!}{
\begin{tabular}{l|c|cccc}\toprule
Sparse Target & Dense & SR-STE	& \mdgf-Exponential	& \sdgf-Stepwise	& \sdgf-Geometric \\\midrule
% \multirow{6}{*}{\makecell{Sparsity\\ Target}}

1:4 (FF)	& 83.45 & 82.355 & \textbf{82.491} & 82.267 & 82.469 \\
1:8  (FF)	& 83.45 & 81.437 & \textbf{81.466} & 81.382 & 81.382 \\
1:16 (FF)	& 83.45 & 80.154 & \textbf{80.542} & 80.386 & 80.274 \\
1:32 (FF)	& 83.45 & 78.972 & \textbf{79.545} & 76.480 & 79.277 \\
1:8   (FF) + 1:8(QK) & 83.45 & 81.441	& \textbf{81.550} & 81.218 & 81.438 \\
% 1:8   (FF) + 1:4(QK) & 76.389 & 78.144	& 78.025 & \textbf{78.273} & 77.163 & 5 \\
% 1:8   (FF) + 1:4(QKV)& 76.389 & 78.222	& 78.319 & \textbf{78.319} & 77.309 & 6 \\
\bottomrule
\end{tabular}
}
\label{table:swin_ff}
\end{table*}

\subsection{Language Understanding \texorpdfstring{$\mapsto$}{} \xx{T5X-Base}}
We also analyze the efficacy of the proposed decaying-based training recipes for the language understanding task. 
We employ a dense pre-trained \xx{T5X-Base} model trained on the C4 dataset with a span-corruption objective~\citep{t5_model}.
The dense pre-trained model undergoes fine-tuning using the GLUE dataset~\citep{wang2019glue} with various training recipes for N:M structured sparsity.

\autoref{table:t5x_ff} depicts the overall score, summarized across eight different GLUE tasks.
We observer a consistent trend where \sdgf outperforms SR-STE at high-sparsity patterns and increasing number of sparse layers.
Notably, we observe a relative difference of $\Delta=+5.3$ in 1:8 ($\mathcal{FF}$) + 1:8 ($\mathcal{QKV}$) sparsity pattern.
\autoref{sec:sup_ablation_studies} and \autoref{sec:sup_t5x_results} provide details about the \xx{T5X-Base} model, per-task evaluation metrics, and additional ablation studies.

\begin{table*}[!th]\centering
\caption{The GLUE overall score on the sparsified \xx{T5X-Base} model across different N:M sparse training recipes and patterns.}
\vspace{0.5em}
\scriptsize
\resizebox{0.8\textwidth}{!}{
\begin{tabular}{l|c|c|ccc}\toprule
Model & Sparsity Target & Dense & SR-STE	& \sdgf-Stepwise	& \sdgf-Geometric \\\midrule
\xx{T5X-Base} & 1:4 (FF)	& 86.2 & \textbf{84.1} & 83.7 ($\Delta=-0.4$) & 83.4 \\
\xx{T5X-Base} & 1:32  (FF) & 86.2 &  79.4 & \textbf{80.9 ($\bm{\Delta=+1.5}$)} & 79.3 \\
\xx{T5X-Base} & 1:8 (FF) + 1:8 (QK)	& 86.2 &  75.8 & \textbf{80.7 ($\bm{\Delta=+4.9}$)} & 76.8\\
\xx{T5X-Base} & 1:8 (FF) + 1:4(QKV) & 86.2 &  78 &  \textbf{80.3 ($\bm{\Delta=+2.3}$)} & 78.9\\
\xx{T5X-Base} & 1:8 (FF) + 1:8 (QKV)	& 86.2 &  74.2 & \textbf{79.5 ($\bm{\Delta=+5.3}$)} & 75.8\\
\bottomrule
\end{tabular}
}
\label{table:t5x_ff}
\end{table*}
\subsection{Language Translation \texorpdfstring{$\mapsto$}{} \xx{Enc-Dec}}
%
% \begin{table*}[!htp]\centering
% \caption{WMT accuracy  using attention-based model with different N:M sparsity targets.\ay{remove unstructured and remove mishra etc.}}
% \scriptsize
% \resizebox{0.98\textwidth}{!}{
% \begin{tabular}{lr|r|cccc|cc}\toprule
% \multicolumn{2}{c|}{Accuracy} &Dense &\multicolumn{4}{c|}{Structure Sparsity} & \multicolumn{2}{c}{Unstructure Sparsity} \\\cmidrule{1-9}
% \multicolumn{2}{c|}{Schedule} &Dense & \sdgf-Stepwise & \mdgf-Linear &Dense Sparse~\citep{mishra2021accelerating} &SR-STE~\citep{sr_ste} &Dense-Sparse &Unstr Sparse \\\midrule
% \multirow{4}{*}{\makecell{Sparsity\\ Target}} &1:16 &0.747 &\textbf{0.717} &\textbf{0.717} &0.714 &0.709 &0.714 &0.714 \\
% &1:32 &0.747 &0.713 &\textbf{0.714} &0.710 &0.707 &0.711 &0.712 \\
% &1:64 &0.747 &0.710 &\textbf{0.711} &0.708 &0.707 &0.711 &0.711 \\
% &1:128 &0.747 &0.708 &\textbf{0.711} &0.708 &0.707 &0.708 &0.709 \\
% \bottomrule
% \end{tabular}
% }
% \label{table:overall}
% \end{table*}
\begin{table*}[!htp]\centering
\caption{The translation accuracy on WMT task across different N:M sparsity patterns and training recipes.}
\vspace{0.5em}
\scriptsize
\resizebox{0.8\textwidth}{!}{
\begin{tabular}{c|c|c|ccc}\toprule
Model&Sparsity Target&Dense&SR-STE&\sdgf-Stepwise&\mdgf-Exponential\\\cmidrule{1-6}
\xx{Enc-Dec (WMT)}&1:16&0.747&0.709&\textbf{0.717}&\textbf{0.717}\\
\xx{Enc-Dec (WMT)}&1:32&0.747&0.707&0.713&\textbf{0.714}\\
\xx{Enc-Dec (WMT)}&1:64&0.747&0.707&0.710&\textbf{0.711}\\
\xx{Enc-Dec (WMT)}&1:128&0.747&0.707&0.708&\textbf{0.711}\\\bottomrule
\end{tabular}
}
\label{table:overall}
\end{table*}
Finally, we compare the performance of different sparse training recipes on WMT language translation task~\citep{wmt_dataset}.
For that, we use an encoder-decoder transformer-based model~\citep{vaswani2017attention} each with six layers and 16 heads, which is relatively smaller than \xx{T5X-Base}.
\label{sec:sup_hyperparams} outlines the details about this model and the training hyperparameters.
%
% \ay{Can you make sure these information are available in appendix and remove it from here -> The embedding dimension for input and Q/K/V is 1024. The FF blocks within each block have 4096 neurons.}
% %
% \ay{Also make sure that we mention which layer we sparsify}
%

\autoref{table:overall} demonstrates the accuracy results across range of sparsity patterns and training recipes.
We observe that \sdgf and \mdgf collectively outperform SR-STE across various N:M structured sparsity patterns.
However, we note that the difference in accuracy achieved through different training recipes is relatively smaller.
This can be attributed to the model size (6 layers vs. 12 layers in \xx{T5X-Base}), as well as the nature of the translation task, which appears to be less sensitive to sparsity patterns and training recipes\footnote{Model accuracy is less affected as we increase the sparsity level beyond 1:32.}.
%
% We compare \mdgf and \sdgf, with  baseline Dense-Sparse~\citep{mishra2021accelerating} and SR-STE~\citep{sr_ste} in \autoref{table:overall}. We evaluate the methods on different sparsity targets.
%
% \autoref{table:overall} shows that Dense-Sparse performs similarly to SR-STE, and \mdgf-Linear achieves the best accuracy across all sparsity targets.
%
% \sdgf-Step performs the second best. More interestingly, \mdgf-Linear can help achieve similar or better accuracy than the ``unstructured sparsity'' ones.
%
% \emph{Our results indicate that the \mdgf pruning method on dense layers enables models to be pruned structurally while achieving comparable or even better accuracy to ``unstructured pruning''.}

\subsection{Baseline Comparison}

\begin{table*}[!htp]\centering
    \caption{Comparing various sparsification techniques by fine-tuning T5X on GLUE dataset. }
    \vspace{0.5em}
    \centering
    \begin{tabular}{c|c|c|c|c}
         Sparse Target & SR-STE \citep{sr_ste} & SNIP \citep{lee2018snip} & IDP \citep{Fang_2022}   & \mdgf-Exponential \\
          % & &  & \citep{} & \\
         \midrule
         1:32 (FF) & 79.4 & 79.5  & 80.6  & \textbf{80.9}\\
    \end{tabular}
    \label{table:baselines}
    \vspace{1em}
\end{table*}

SR-STE is our primary baseline in our evaluations as it has shown good results in low-sparsity regions [2:4,1:4] and is considered SOTA for N:M training. 
We also compared against other techniques like Inherited Dynamic Pruning (IDP)~\citep{Fang_2022}, and SNIP: Single-shot Network Pruning~\citep{lee2018snip}. \autoref{table:baselines} compares the results on T5X with GLUE dataset.
We also tried to test against LBC~\citep{zhang2022learning} but could not recreate the results shown in the paper.\footnote{We have contacted the authors but cannot solve the issue.}

\subsection{Recipe impact for CNNs.}
While the primary focus of this work is on evaluating sparse training recipe for transformer models, for the sake of completeness, we also test the efficacy of our recipe on CNNs.
We train \xx{ResNet-50} following two sparse training recipes (SR-STE and \mdgf-Exponential) and across different sparse patterns (2:8, 1:8).
We pruned all the convolution layers and evaluate Top-1 validation accuracy on CIFAR-10. 
\autoref{table:resnet50} shows a similar pattern, decaying-based sparse training recipes outperform SR-STE in both cases.
\begin{table}[!htp]
    \centering
    \caption{ResNet-50 Top-1 validation accuracy.}
    % \resizebox{0.6\textwidth}{!}{
    \begin{tabular}{c|c|ccc}
    \toprule
    Sparse Target	& Dense	 & SR-STE & \mdgf-Exponential \\
    \midrule
        2:8 & 85.09 & 83.33 & \textbf{83.60}\\
        1:8 & 85.09 & 80.78 & \textbf{82.48}\\
    \bottomrule
    \end{tabular}
    % }
    \label{table:resnet50}
\end{table}

\section{Limitations and Future Works} 
The prevalence of self-attention models and their growing parameter size inspired this work.
The primary objective of this research is to enable effective sparsity (acceptable quality) with high ratio for such models. 
While in this paper, we evaluate the proposed sparse training recipes in isolation (either \mdgf or \sdgf), combining these methods at different training region can potentially lead to higher model quality.
The main finding of our work is that pruning in the high sparsity regime adversely affects gradient estimation, consequently resulting in suboptimal model quality.
To mitigate this undesired phenomenon, we propose a strategy of progressively tightening the gradient flow for pruned weights.
Our results show that this idea, while simple, proves to be effective across a variety of models and datasets.
%
% In addition, we only study one combination of sparsification recipe (Either \mdgf or \sdgf. We can potential combined \sdgf and \mdgf, like \sdgf-Geometric + \mdgf-Exponential at the same time, and so on. We might discover better recipe by exploring some other combinations such as salient-based pruning + \mdgf-linear + N:M structured pruning, magnitude pruning + \sdgf-stepwise + unstructured sparsity, or many others.

% 3) Lastly, in the evaluations, most of the hyper-parameters are taken for which performs best for dense baselines. We did a limited scope of hyper-parameters sweep in our ablations studies. A full-fledged hyper-parameter search might discover more performance improvement in \mdgf and \sdgf. 4) The \mdgf training cost during sparsify steps would be higher than SR-STE, as it has a non-binary mask for which training time would be equivalent to dense training. 5) Domino Search\citep{sun2021dominosearch} explores mixed N:M sparsity schemes for pre-trained dense neural networks, surpassing the accuracy of uniform-sparsity schemes within equivalent complexity constraints. Similar to them, we can experiment with inducing different amounts of sparsity in different layers of the model.

\section{Conclusion}
This work studies the efficacy of recent sparsity recipes for N:M sparsity across range of transformer-based models.
We observe that conventional methods introduce nontrivial noise to gradient estimates, particularly at high-sparsity regimes (>75$\%$).
Building on this observation, we propose and compare a class of decaying-based training recipes for N:M structured sparsity.
Our results demonstrate that our recipe, \mdgf-Exponential, consistently deliver SOTA model accuracy for a variety of vision and language models, with more than $\sim$2\% (vision) and $\sim$5\% (language) improvement at high sparsity regime.
We empirically show that the effectiveness of the proposed recipes primarily depending on the gradient flow, especially at the initial training steps.
Finally, we compare the sparse training recipes in terms training and inference FLOPs.
At iso-FLOPs for training, our approach offers 2\% higher accuracy.
In addition, we demonstrate that \mdgf-Exponential (1:16) yields comparable accuracy to SR-STE (2:4), resulting in approximately 60\% fewer inference FLOPs and 30\% fewer parameters. 
%
% Our implementation code-base can be found at  \url{https://anonymous.4open.science/r/n_m_decay_1605-496B}. 
% Finally, we associate each of these training recipes with training compute cost (FLOPs) and study the trade-offs between model accuracy and compute cost.
%
% The results demonstrate that both proposed methods deliver better model quality at iso-training cost.

\section*{Acknowledgements}
We would like to extend our gratitude towards Jeremiah Willcock, Penporn Koanantakool, Chandu Thekkath, Cliff Young, and James Laudon for their invaluable feedback on the early draft of this work.
We also appreciate the support from our extended team at Google DeepMind.

\bibliography{references}
\bibliographystyle{icml2024}

%%%%%%%%%%%%%%%%%%%%%%%%%%%%%%%%%%%%%%%%%%%%%%%%%%%%%%%%%%%%%%%%%%%%%%%%%%%%%%%
%%%%%%%%%%%%%%%%%%%%%%%%%%%%%%%%%%%%%%%%%%%%%%%%%%%%%%%%%%%%%%%%%%%%%%%%%%%%%%%
% APPENDIX
%%%%%%%%%%%%%%%%%%%%%%%%%%%%%%%%%%%%%%%%%%%%%%%%%%%%%%%%%%%%%%%%%%%%%%%%%%%%%%%
%%%%%%%%%%%%%%%%%%%%%%%%%%%%%%%%%%%%%%%%%%%%%%%%%%%%%%%%%%%%%%%%%%%%%%%%%%%%%%%
\newpage
\appendix
\onecolumn

% \newpage
% \appendix
% \addcontentsline{toc}{section}{Appendix} % Add the appendix text to the document TOC
% \renewcommand{\thesection}{\Alph{section}}
% \setcounter{section}{0}
% \part{Appendix} % Start the appendix part
% \parttoc % Insert the appendix TOC
\section{Ablations Studies}
\label{sec:sup_ablation_studies}
This section shows the various ablation studies we performed during our experiments. 

% \subsection{Dense training v.s. Training from scratch for SR-STE.}
% \label{sec:ab_srste}
% %
% SR-STE uses sparse training from scratch. All the other methods we evaluated have a dense training phase at the first few steps (epochs).
% %
% This recipe has been proven to be effective as shown in the previous experiments and many prior works~\citep{mishra2021accelerating,pool2021channel, frankle2018lottery,lee2018snip,evci2019difficulty,han2015deep, guo2016dynamic,he2017channel,molchanov2016pruning,yao2019balanced,zhu2017prune,gamboa2020campfire,narang2017exploring,narang2017block,elsen2020fast,park2018squantizer,kalchbrenner2018efficient,evci2020rigging}.
% %
% Therefore, we experiment on adding a dense training phase at the beginning of SR-STE training, as shown in \autoref{table:ab_srste}.
% %
% We found that adding few steps of dense training (1.25\% - 10\% of the total training steps) can increase the accuracy by around 0.002 to 0.003. This tells that few steps of dense training does help achieve better performance even for SR-STE. Interestingly, the improved SR-STE becomes competitive to the proposed Structure Decay.
% %
% However, Mask Decay is still consistently better.

% \input{Tables/ablation_srste}

\subsection{Effect of dense training steps \texorpdfstring{($d$)} 
.}
\label{sec:ab_d_steps}
Both our proposed methods, \mdgf and \sdgf include a dense training phase. We do an ablation study on different amounts of dense training steps(\% of total steps) in \autoref{table:ab_d_steps}.
We perform this study on the language translation model (more implementation details in section \autoref{sec:langugae_model_impl}) trained on WMT-17.
We found that changing the dense step between 1.25\% - 10\% of the total training steps does not observably change the accuracy performance.
However, empirically, we found that the dense training phase is still essential. The model cannot achieve as competitive accuracy without few epochs of dense training. 

\begin{table*}[!htp]\centering
\caption{Ablation: The effect of number of dense training steps ($d$).}
\scriptsize
\begin{tabular}{lrrrrrrrrrr}\toprule
\multicolumn{2}{c}{Accuracy} &\multicolumn{4}{c}{\mdgf-Linear} &\multicolumn{4}{c}{\sdgf-Stepwise} \\\cmidrule{1-10}
\multicolumn{2}{c}{Sparsity Target} &1:16 &1:32 &1:64 &1:128 &1:16 &1:32 &1:64 &1:128 \\\midrule
\multirow{4}{*}{Dense steps (d)} & 1.25\% &0.7155 &0.7134 &0.7106 &0.7100 &0.7157 &0.7134 &0.7108 &0.7106 \\
& 2.5\% &\textbf{0.7160} &0.7127 &\textbf{0.7110} &0.7093 &0.7160 &0.7136 &\textbf{0.7117} &0.7100 \\
& 5\% &0.7157 &\textbf{0.7137} &0.7103 &0.7094 &0.7164 &\textbf{0.7141} &0.7107 &0.7098 \\
& 10\% &0.7156 &0.7126 &0.7107 &\textbf{0.7104} &\textbf{0.7165} &0.7128 &0.7115 &\textbf{0.7107} \\
\bottomrule
\end{tabular}
\label{table:ab_d_steps}
\end{table*}

\subsection{Effects of fine-tuning steps \texorpdfstring{($s$)}.}
\label{sec:ab_s_steps}
We also have a sets of study on number of fine-tuning steps in \autoref{table:ab_s_steps}.
We perform this study on the language translation model (more implementation details in section \autoref{sec:langugae_model_impl}) trained on WMT-17.
We found that for all of our proposed methods, the fine-tuning steps between 10\% - 20\% of the total training steps do not observably change the accuracy performance.
However, empirically, we also found few steps of fine-tuning at the end are essential to recovering the accuracy. 
\begin{table*}[!htp]\centering
\caption{Ablation: The effect of number of fine-tuning steps ($s$).}
\scriptsize
\begin{tabular}{lrrrrrrrrrr}\toprule
\multicolumn{2}{c}{Accuracy} &\multicolumn{4}{c}{\mdgf-Linear} &\multicolumn{4}{c}{\sdgf-Stepwise} \\\cmidrule{1-10}
\multicolumn{2}{c}{Sparsity Target} &1:16 &1:32 &1:64 &1:128 &1:16 &1:32 &1:64 &1:128 \\\midrule
\multirow{2}{*}{Fine-tuning steps (s)} & 10\% &0.7153 &0.7130 &\textbf{0.7107} &\textbf{0.7098} &\textbf{0.7160} &\textbf{0.7125} &\textbf{0.7095} &\textbf{0.7072} \\
&20\% &\textbf{0.7161} &\textbf{0.7132} &0.7106 &0.7097 &0.7121 &0.7093 &0.7081 &0.7065 \\
\bottomrule
\end{tabular}
\label{table:ab_s_steps}
\end{table*}

\subsection{Effect of \texorpdfstring{($\beta^t$) in \mdgf-Linear}\ }
We also study on effect of decay rate on model's accuracy in \autoref{table:linear_decay_ablation}.
We do experiments with varying $\beta^t$ for ViT-Base trained on Imagenet-1k for different sparsity targets.

We observe that a higher decay rate is beneficial at low sparsity targets (2:4,1:4), but for targets higher than 1:8, we found lower decay rate works better.

\begin{table}[!ht]\centering
\vspace{-1em}
\caption{Ablation: The effect of mask decay rate ($\beta^t$) for \mdgf-Linear.}
\scriptsize
\begin{tabular}{lr|rrr}\toprule
% \multicolumn{2}{c}{Accuracy} &\multicolumn{4}{c}{Mask Decay} \\\cmidrule{1-6}
\multicolumn{2}{c}{Sparsity Target} &2:4 &1:4 &1:8 \\\midrule
\multirow{2}{*}{Mask decay rate ($\beta^t$)} &0.0002 & 77.495 &78.448 & \textbf{78.019}  \\
&0.001 &\textbf{77.613} & \textbf{78.512} &76.4075 \\
\bottomrule
\end{tabular}
\label{table:linear_decay_ablation}
\end{table}
\section{Detailed Results for \xx{T5X-Base} Sparsification on GLUE Dataset}
\label{sec:sup_t5x_results}

We compared sparsification methods N:M block sparsification against state-of-the-art technique, SR-STE on.  T5 model uses a span-based masked language modeling (MLM) objective. T5 models were introduced in \cite{t5_model} and the updated models are available at \href{https://github.com/google-research/t5x}{T5X-github}. We train a pre trained t5x-base model on GLUE dataset \citep{wang2019glue}.

The main paper shows a snapshot of the performance across various sparsity targets using the overall score as metric. \autoref{table:t5x_full} presents all 9 scores for each sparsification technique and sparsity target. 

\begin{table*}[!htp]\centering
\caption{GLUE full score using various T5X-base with different N:M sparse targets and various sparsification techniques.}
\scriptsize
\resizebox{\textwidth}{!}{
\begin{tabular}{l|c|cccccccccc}\toprule
&  & overall score &	CoLA &	MNLI matched & MNLI mismatched & MRPC & QNLI & QQP & RTE & SST-2 & STS-B \\
\midrule
Dense &                  -                        & 86.2 & 58.9 & 87.2 & 87 & 92.4 / 89.2 (90.8) &93.6 & 92.0 / 89.2 (90.6) & 	82.3 & 95 & 90.1 / 90.0 (90.0) \\ \midrule
SR-STE (Zero Dense) & 	1:4             	      & 83.1	& 41.8	& 85.2	& 85.3	& \textbf{92.8} / \textbf{90.0 (91.4)}	& \textbf{92.3}	& 91.8 / 88.9 (90.3)	& 79.1	& \textbf{93.6}	& \textbf{89.5 / 89.2 (89.3)} \\
SR-STE (10K Dense) & 	1:4                       & \textbf{84.1}	&  48.1 & \textbf{85.7} & \textbf{85.6} &	92.4 / 89.5 (91.0) &	92.1 &	\textbf{91.8 / 89.0 (90.4)} & \textbf{82.}7 &	\textbf{93.6} &	87.9 / 87.7 (87.8) \\
\mdgf-Stepwise (10K Dense) & 	1:4	              & 83.7	& \textbf{48.8}	& 85.3	& 85.4	& 92.4 / 89.2 (90.8)	& 92.3	& 91.8 / 89.0 (90.4)	& 80.5	& 93.5	& 86.5 / 86.3 (86.4) \\
\mdgf-Geometric (Zero Dense) &	1:4	              & 83.3	& 48.4	& 85.3	& 85.3	& 92.0 / 89.0 (90.5)	& 91.8	& 91.8 / 88.9 (90.3)	& 78	& 92.8	& 87.3 / 87.4 (87.3) \\
\mdgf-Geometric (10K Dense) & 	1:4               & 83.4	&  47.2 & 85.4 & 85.3 &	92.6 / 89.7 (91.1) &	92	 &  \textbf{91.8 / 89.0 (90.4}) & 79.8 &	92.9 &	86.7 / 86.4 (86.5) \\
\midrule
SR-STE (Zero Dense) & 	1:32	                   & 77.1	& 19	& 81.3	& 81.3	& 90.9 / 87.0 (89.0)	& 86.9	& 90.6 / 87.4 (89.0)	& 71.1	& 89.9	& 86.7 / 86.8 (86.8) \\
SR-STE (10K Dense) & 	1:32	                   & 79.4	&  29.4 & 82.2 & 82.6 &	91.5 / 88.5 (90.0) &	89.6 &	91.2 / 88.2 (89.7) & 72.6 &	\textbf{91.4} &	\textbf{87.1 / 87.2 (87.2}) \\
\mdgf-Stepwise (10K Dense) & 	1:32	           & \textbf{80.9}	& \textbf{38.3}	& \textbf{83.6}	& \textbf{83.7}	& \textbf{92.5} / \textbf{89.7 (91.1)}	& \textbf{90.5}	& \textbf{91.5 / 88.5 (90.0)}	& \textbf{74.4}	& 91.2	& 85.2 / 85.0 (85.1) \\
\mdgf-Geometric (Zero Dense) &	1:32	           & 77.6	& 20.2	& 81.3	& 81.6	& 91.8 / 88.5 (90.1)	& 87.2	& 90.8 / 87.7 (89.2)	& 73.3	& 90.1	& 85.8 / 85.5 (85.6) \\
\mdgf-Geometric (10K Dense) & 	1:32	           & 79.3	&  29.2 & 82.3 & 82.9 &	91.3 / 88.0 (89.6) &	90.4 &	91.3 / 88.3 (89.8) & 73.3 &	90.5 &	85.4 / 85.4 (85.4) \\

\midrule
SR-STE (Zero Dense) & 	1:8(FF) + 1:8(QK)	        & 74.4	& 15.7	& 77.2	& 77.6	& 89.9 / 85.8 (87.8)	& 83.6	& 89.7 / 86.2 (87.9)	& 67.5	& 88.2 & 84.1 / 83.9 (84.0) \\
SR-STE (10K Dense) & 	1:8(FF) + 1:8(QK)	        &  75.8	&  19.9 & 78.6 & 79.4 &	89.7 / 86.0 (87.9) &	84	 &  90.1 / 86.7 (88.4) & 70	  &   89.4 &	84.5 / 84.2 (84.4) \\
\mdgf-Stepwise (10K Dense) & 	1:8(FF) + 1:8(QK)   & \textbf{80.7}	& \textbf{38.7}	& \textbf{83.1}	& \textbf{83.2}	& \textbf{90.9 / 87.7 (89.3)}	& \textbf{89.9}	& \textbf{91.2 / 88.2 (89.7)}	& \textbf{76.2}	& \textbf{91.9}	& \textbf{84.5 / 84.5 (84.5)} \\
\mdgf-Geometric (Zero Dense) &	1:8(FF) + 1:8(QK)   &  75.8	&  21.6	& 78.8	& 79	    & 90.0 / 86.0 (88.0)	& 83.6	& 90.1 / 86.6 (88.3)	& 69.7	& 88.9	& 84.0 / 83.9 (83.9) \\
\mdgf-Geometric (10K Dense) & 	1:8(FF) + 1:8(QK)   &  76.8	&  22.3 & 80.7 & 80.9 &	89.8 / 85.8 (87.8) &	86.3 &	90.5 / 87.4 (89.0) & 70	  &   91.1 &	83.7 / 83.4 (83.6) \\
\midrule
SR-STE (Zero Dense) &   1:8(FF) + 1:8(QKV)	        & 73.2	& 13.5	& 76.3	& 76.4	& 89.0 / 84.6 (86.8)	& 83.2	& 89.5 / 85.9 (87.7)	& 63.9	& 87	 &   84.3 / 84.2 (84.2) \\
SR-STE (10K Dense) & 	1:8(FF) + 1:8(QKV)	        &  74.2	&  16.1 & 77.7 & 77.6 &	88.5 / 84.1 (86.3) &	82.9 &	89.9 / 86.3 (88.1) & 66.4 &	88.8 &	84.4 / 84.2 (84.3)\\
\mdgf-Stepwise (10K Dense) & 	1:8(FF) + 1:8(QKV)  & \textbf{79.5}	& \textbf{33}	& \textbf{82.3}	& \textbf{82.3}	& \textbf{91.3 / 87.7 (89.5)}	& \textbf{89.2}	& \textbf{91.0 / 88.0 (89.5)}	&\textbf{ 74.4}	& \textbf{91.1}	& \textbf{84.5 / 84.8 (84.6)} \\
\mdgf-Geometric (Zero Dense) &	1:8(FF) + 1:8(QKV)  &  75.5	&  22.1	& 78.6	& 78.7	& 90.5 / 86.8 (88.6)	& 83.4	& 90.0 / 86.5 (88.2)	& 67.9	& 88.2	& 84.2 / 84.2 (84.2) \\
\mdgf-Geometric (10K Dense) & 	1:8(FF) + 1:8(QKV)  &  75.8	&  19.5 & 79.4 & 79.6 &	89.4 / 85.3 (87.3) &	84.5 &	90.2 / 86.8 (88.5) & 70.4 &	89.8 &	83.3 / 83.0 (83.2) \\
\midrule
SR-STE (Zero Dense) & 	1:8(FF) + 1:4(QKV)	        & 75.1	& 15	& 78.4	& 79	&     90.5 / 86.8 (88.6)& 	84.2& 	90.1 / 86.6 (88.4)	& 67.9	& 88.4	& \textbf{86.2 / 86.1 (86.2)} \\
SR-STE (10K Dense) & 	1:8(FF) + 1:4(QKV)	        &  78	&  24.5 & 81.2  & 81.6  &	91.1 / 87.7 (89.4) &	87.1 &	90.6 / 87.3 (89.0) & 72.2 &	90.9 &	85.8 / 85.8 (85.8) \\
\mdgf-Stepwise (10K Dense) & 	1:8(FF) + 1:4(QKV)  & \textbf{80.3}	& \textbf{36.4}	& \textbf{83.2}	& \textbf{83.4}	& 90.9 / 87.3 (89.1)	& \textbf{90.3}	& \textbf{91.3 / 88.3 (89.8)}	& \textbf{74.7}	& 90.9	& 85.2 / 85.0 (85.1) \\
\mdgf-Geometric (Zero Dense) &	1:8(FF) + 1:4(QKV)  &  76.8	&  20.2	& 80.5	& 80.8	& \textbf{91.3 / 87.7 (89.5)}	& 85.4	& 90.3 / 87.0 (88.6)	& 70.8	& 90.4	& 84.9 / 84.9 (84.9) \\
\mdgf-Geometric (10K Dense) & 	1:8(FF) + 1:4(QKV)  &  78.9	&  27.7 & 82.4  & 82.4  &	\textbf{91.3 / 87.7 (89.5)} &	88.8 &	91.0 / 88.1 (89.6) & 74.4 &	\textbf{91.3} &	84.5 / 84.5 (84.5) \\
\bottomrule
\end{tabular}
}
\label{table:t5x_full}
\end{table*}

Here is an itemized list of nine tasks used in the GLUE dataset, along with brief descriptions of each:
\begin{itemize}
    \item \niparagraph{CoLA (Corpus of Linguistic Acceptability)}: Classify whether a given sentence is grammatically acceptable or not.

    \item \niparagraph{MNLI (Multi-Genre Natural Language Inference)}: Classify the relationship between a given premise and hypothesis as entailment, contradiction, or neutral. We use the standard test set, for which we obtained private labels from the authors, and evaluate on both the matched (in-domain) and mismatched (cross-domain) sections.
    
    \item \niparagraph{MRPC (Microsoft Research Paraphrase Corpus)}: Determine whether a pair of sentences express the same meaning or not.

    \item \niparagraph{QNLI (Question-answering Natural Language Inference)}: Determine whether a given question can be answered correctly using a given sentence.
    
    \item \niparagraph{QQP (Quora Question Pairs)}: Determine whether a pair of questions from Quora are semantically equivalent or not.
    
    \item \niparagraph{RTE (Recognizing Textual Entailment)}: Classify the relationship between a given premise and hypothesis as entailment or not.
    
    \item \niparagraph{SST-2 (Stanford Sentiment Treebank)}: Determine the sentiment of a given sentence as either positive or negative.
    
    \item \niparagraph{STS-B (Semantic Textual Similarity Benchmark)}: Calculate the similarity score between two sentences on a scale from 0 to 5.

    % \item WNLI (Winograd NLI): Determine if a pronoun in a given sentence has the correct antecedent based on world knowledge.
\end{itemize}
    
These tasks cover various aspects of language understanding, including sentence acceptability, sentiment analysis, paraphrase detection, textual similarity, natural language inference, question-answering, and co-reference resolution.

\begin{figure*}[!hbt]
   % \centering
%
\subfigure{\label{eval_t5x:cola}
\includegraphics[width=0.31\linewidth]{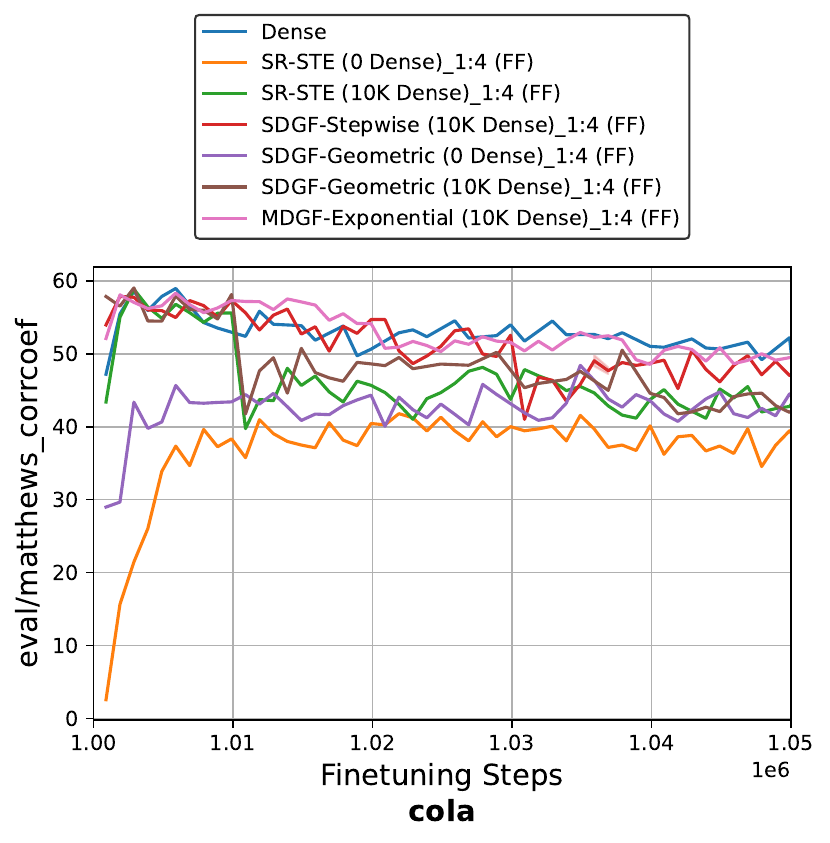}}
\subfigure{\label{eval_t5x:mnli_matched}
\includegraphics[width=0.31\linewidth]{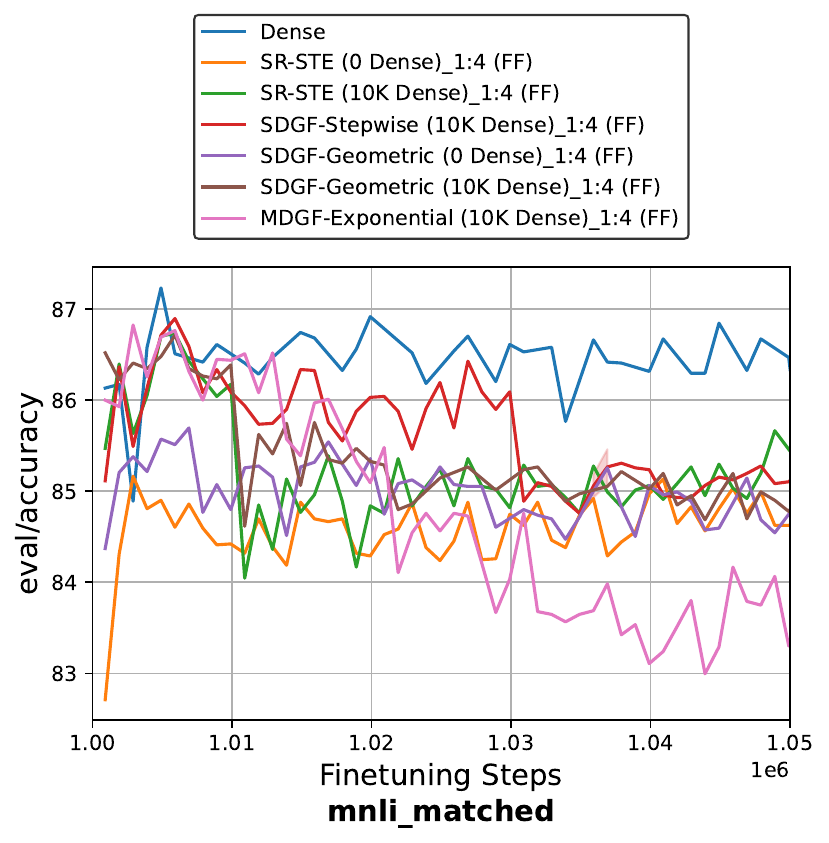}}
\subfigure{\label{eval_t5x:mnli_mismatched}
\includegraphics[width=0.31\linewidth]{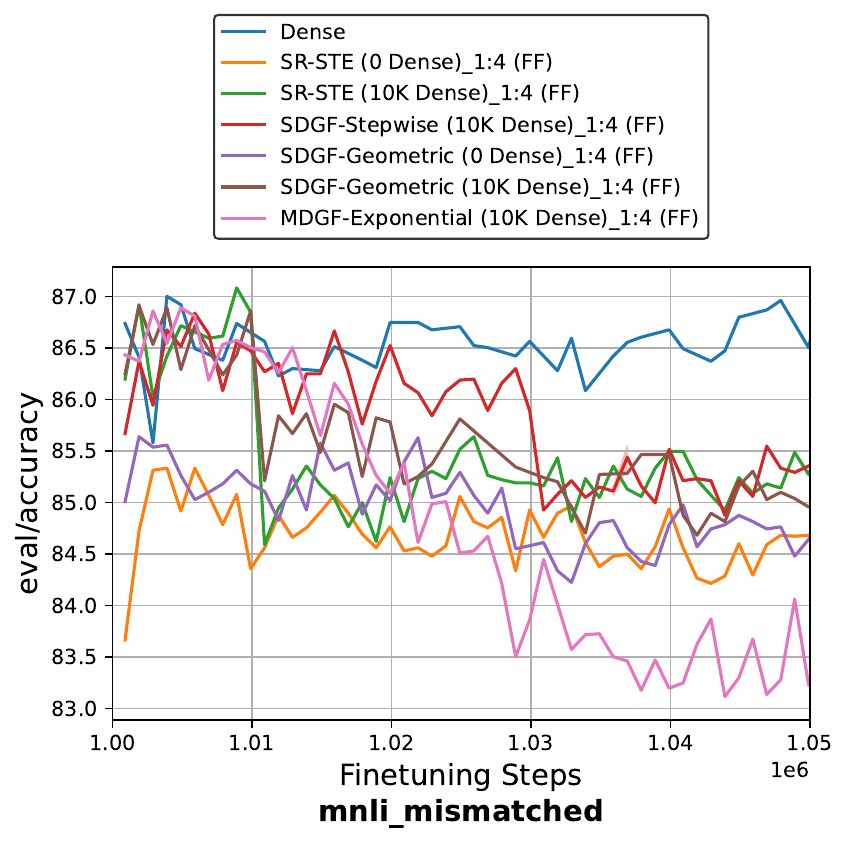}}\\
\subfigure{\label{eval_t5x:mrpc:accuracy}
\includegraphics[width=0.31\linewidth]{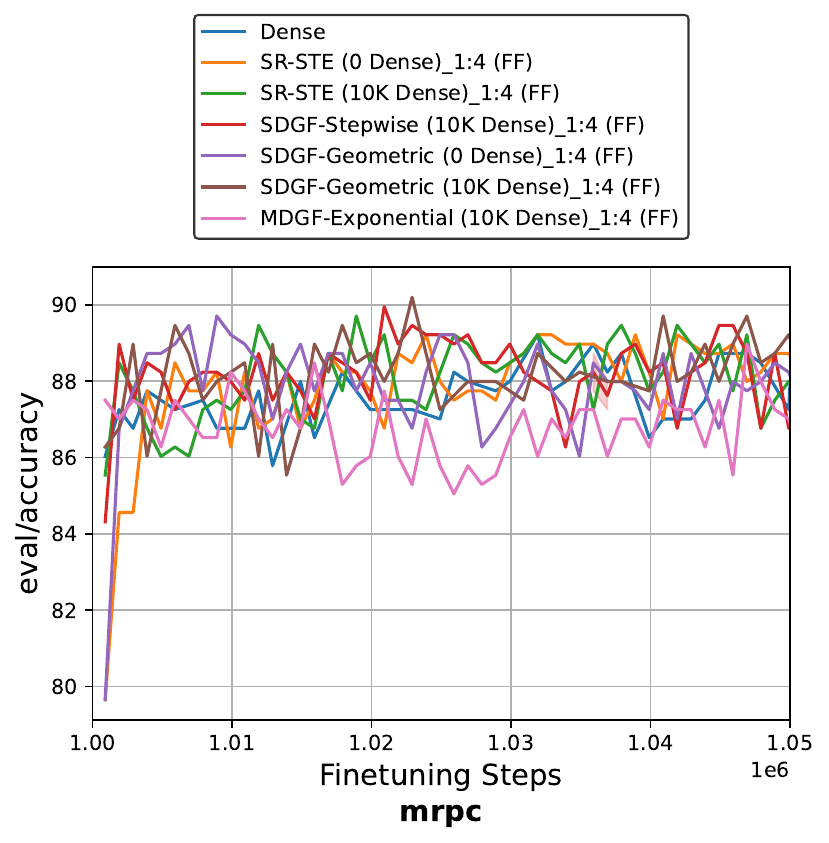}}
\subfigure{\label{eval_t5x:mrpc:f1}
\includegraphics[width=0.31\linewidth]{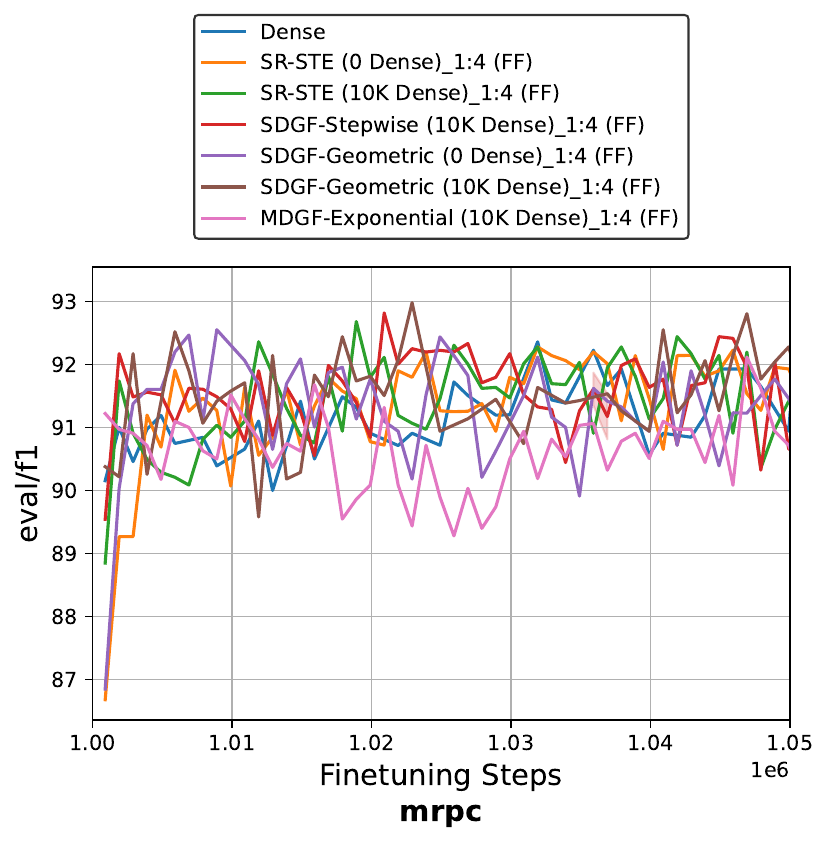}}
\subfigure{\label{eval_t5x:qnli}
\includegraphics[width=0.31\linewidth]{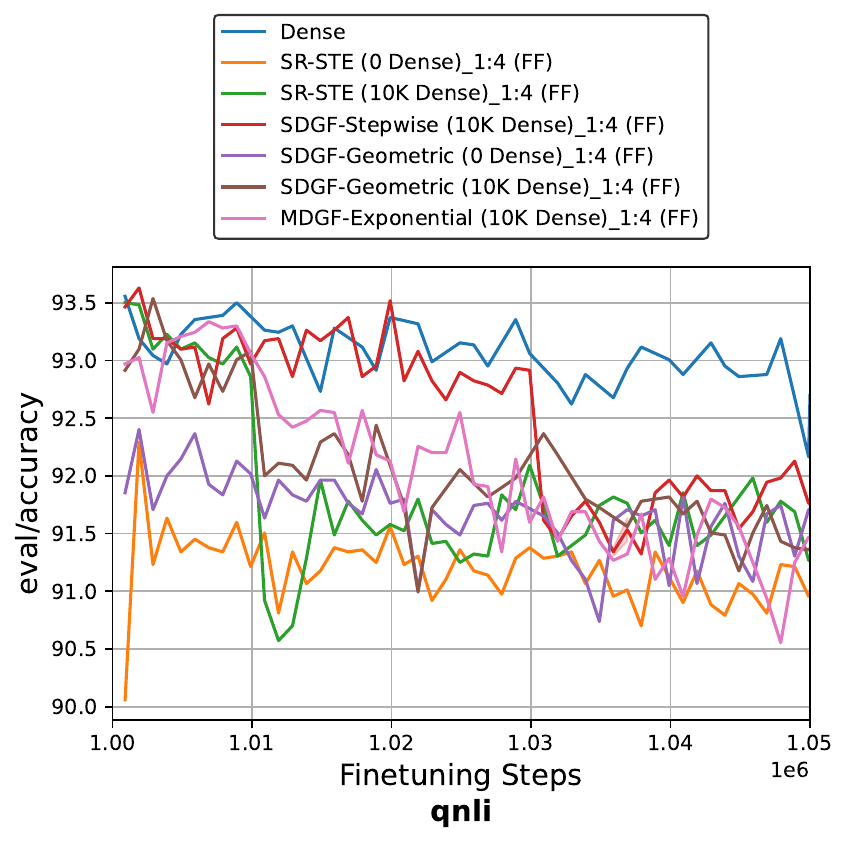}}\\
%
% \subfigure{\label{eval_t5x:qqp:accuracy}
% \includegraphics[width=0.31\linewidth]{eval_t5x_plots_1_4_ff/glue_qqp_v002_eval_accuracy_1_4_ff.pdf}}
%
% \subfigure{\label{eval_t5x:qqp:f1}
% \includegraphics[width=0.31\linewidth]{eval_t5x_plots_1_4_ff/glue_qqp_v002_eval_f1_1_4_ff.pdf}}
% %
% \subfigure{\label{eval_t5x:rte}
% \includegraphics[width=0.31\linewidth]{eval_t5x_plots_1_4_ff/glue_rte_v002_eval_accuracy_1_4_ff.pdf}}\\
% %
% \subfigure{\label{eval_t5x:sst2}
% \includegraphics[width=0.31\linewidth]{eval_t5x_plots_1_4_ff/glue_sst2_v002_eval_accuracy_1_4_ff.pdf}}
% %
% \subfigure{\label{eval_t5x:stsb:pearson_corr}
% \includegraphics[width=0.31\linewidth]{eval_t5x_plots_1_4_ff/glue_stsb_v002_eval_pearson_corrcoef_1_4_ff.pdf}}
% %
% \subfigure{\label{eval_t5x:stsb:spearman}
% \includegraphics[width=0.31\linewidth]{eval_t5x_plots_1_4_ff/glue_stsb_v002_eval_spearman_corrcoef_1_4_ff.pdf}}
\caption{Per-task evaluations for \xx{T5X-Base} model finetuned on the GLUE dataset for 50\,K steps. }
\label{fig:eval_t5x_1_4_ff}
\end{figure*}

% \autoref{fig:cross_ent_loss_t5x} shows the cross entropy loss function for different sparsity targets for all sparsification techniques. We observe that in most cases loss for \sdgf converges faster than SR-STE and hence it performs better than SR-STE. 
% 
\autoref{fig:eval_t5x_1_4_ff} shows the accuracy vs. fine-tuneing step curve for each of the 9 benchmarks of GLUE.

\section{Detailed Experimental Settings}
\label{sec:sup_hyperparams_appx}

\subsection{Datasets}
\subsubsection{ImageNet-1K}
ImageNet-1K \citep{imagenet} is a large-scale image classification task, known as one of the most
challenging image classification benchmarks. It consists of more than 1.2 million training images and
50K validation images with a size of 224x224 pixels, each with 3 channels. Each image is labeled as one of the 1K classes. We use this dataset for studies in Section 4.1 of the main paper. For ViT and SwinV2 experiments, we use a patch size of 16. This converts the 224x224 pixel image into an input of sequence length $224/16 * 224/16  = 196$. 

\niparagraph{Evaluation metrics.} All reported results follow standard Top-1 validation accuracy.
 
\subsubsection{CIFAR10}
CIFAR-10~\citep{cifar10} is a smaller-scale image classification dataset consisting of 10 classes. Each class has
6000 color images of 32x32 pixels in size. 

\niparagraph{Evaluation metrics.} All reported results to follow standard Top-1 accuracy.

\subsubsection{GLUE}
The General Language Understanding Evaluation (GLUE) \citep{wang2019glue} benchmark is a collection of resources for training, evaluating, and analyzing natural language understanding systems. GLUE consists of:
A benchmark of nine sentence- or sentence-pair language understanding tasks built on established existing datasets and selected to cover a diverse range of dataset sizes, text genres, and degrees of difficulty,
\autoref{table:t5x_full} shows the overall score for each sparsity target using different sparsification methods.

\niparagraph{Evaluation metrics.} All reported results in the main paper use the overall average score.

\subsubsection{WMT}
WMT-17 (English-German) \citep{wmt_dataset} is a key benchmark in machine translation research. They hold several translation datasets across different languages. The training set consists of about 4.5 million bilingual
sentence pairs from WMT 2014.  

\niparagraph{Evaluation metrics.} We calculate accuracy by comparing the translated output to the correct translation in the validation datasets. 

\subsection{Hyperparameters for Different Models}
\label{appx:sub:hp}
\subsubsection{Image Classification \texorpdfstring{$\rightarrow$}{} Vision Transformers~(\xx{ViT})}
We train the ViT-Base model on ImageNet-1k with hyperparameters presented in \autoref{table:vit_hyper}. We follow the hyperparameter setting in~\citep{rw2019timm} for all ViT experiments. 
We also use the same hyperparameters to train ViT-Tiny model ( 3 layers, 3 attention head per layer, Embedding dimension: 192) on CIFAR-10 for initial experiments in Section 3.2 for analysing the trends of weights, gradients and optimizer moments and comparing those with SR-STE.

\begin{table}[!ht]\centering
\caption{Hyperparameters used for training ViT on ImageNet-1K.}
\begin{tabular}{lrr}\toprule
Batch Size & 256 \\
Training Epoches & 350 \\

Learning Rate & 1e-3 \\
LR Warmup Epoches & 15 \\
LR Decay schedular & Cosine \\

Decay Rate & 0.1 \\
Decay Epoches & 100 \\

Optimizer & AdamW  \\ 
Optimizer coefs &   beta1 = 0.9, beta2 = 0.999 \\
\bottomrule
\end{tabular}
\label{table:vit_hyper}
\end{table}

The detailed list of all hyperparameters can be found at \href{https://anonymous.4open.science/r/n_m_decay_1605-E77F/vit_base_training.yaml}{hyperparaters.yaml}. For ViT-Base, the training phase takes $\approx$ 44 hours on 16 - A100  GPUs.

\begin{figure*}[!hbt]
\centering
\subfigure[1:8 FF (Top-1 Accuracy)]{\label{fig:vit:acc:1_8_ff}
\includegraphics[width=0.32\linewidth]{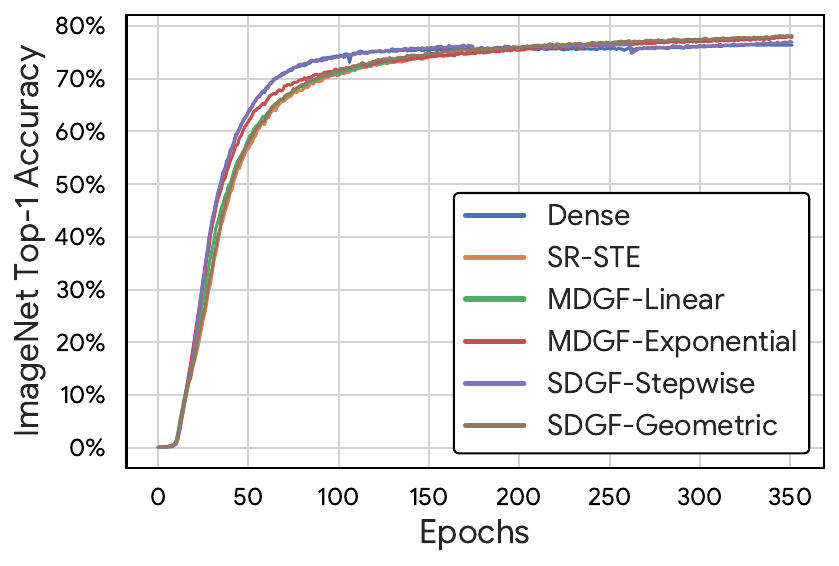}}
\subfigure[1:32 FF (Top-1 Accuracy)]{\label{fig:vit:acc:1_32_ff}
\includegraphics[width=0.32\linewidth]{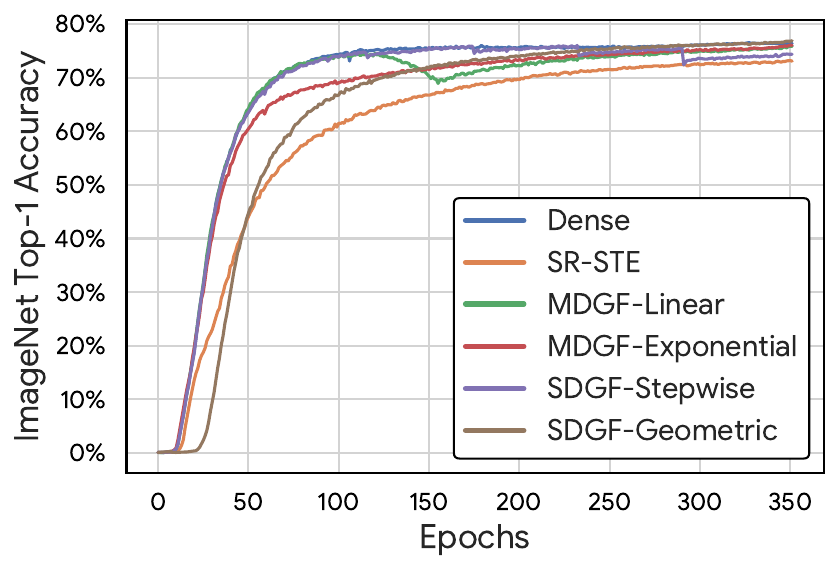}}
\subfigure[1:8 FF+QK (Top-1 Accuracy)]{\label{fig:vit:acc:1_8_ffqk}
\includegraphics[width=0.32\linewidth]{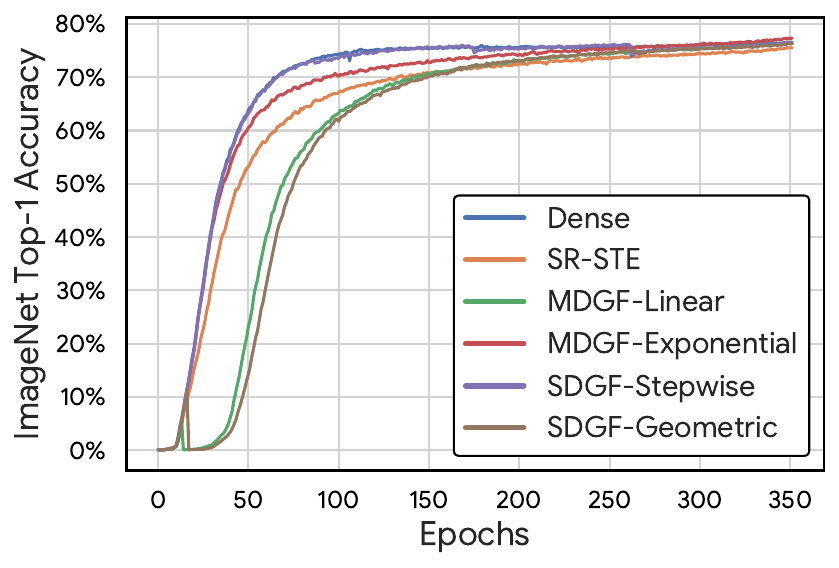}}\\
\subfigure[1:8 FF (Top-5 Accuracy)]{\label{fig:vit:acc:1_8_ff:top5}
\includegraphics[width=0.32\linewidth]{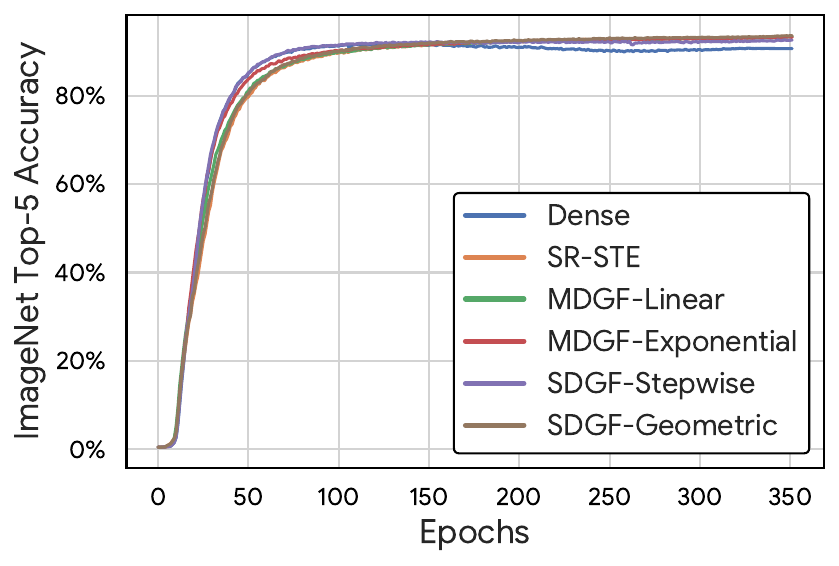}}
\subfigure[1:32 FF (Top-5 Accuracy)]{\label{fig:vit:acc:1_32_ff:top5}
\includegraphics[width=0.32\linewidth]{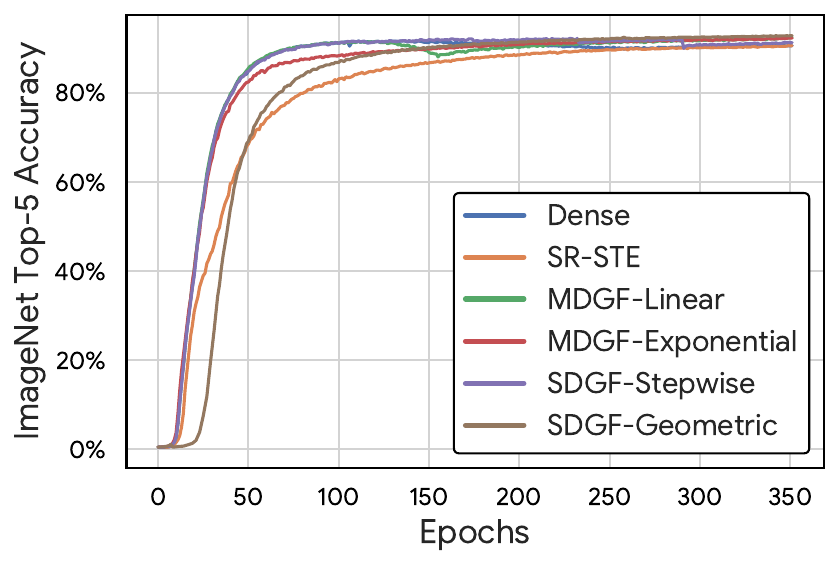}}
\subfigure[1:8 FF+QK (Top-5 Accuracy)]{\label{fig:vit:acc:1_8_ffqk:top5}
\includegraphics[width=0.32\linewidth]{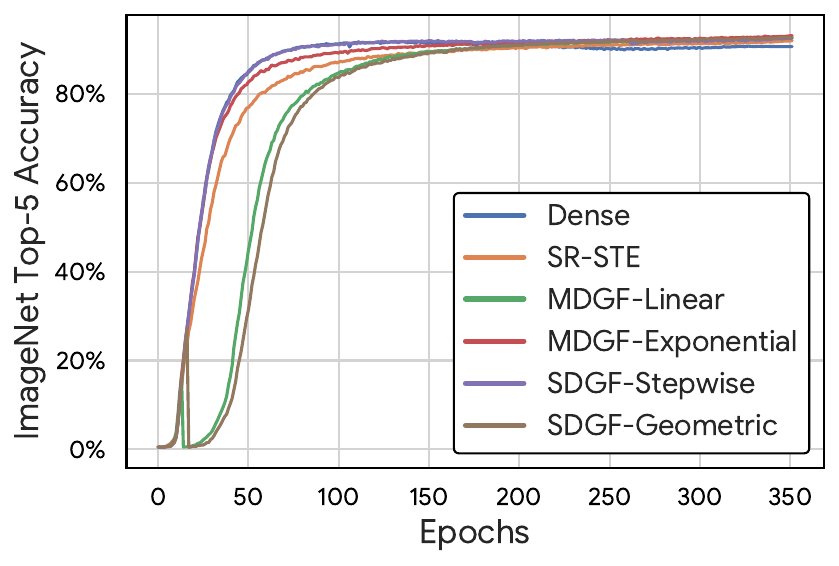}}
\caption{Training Epochs vs Accuracy graph for different sparsity targets. We train \xx{ViT-Base} on ImageNet-1K.}
\label{fig:vit:acc}
\end{figure*}

\autoref{fig:vit:acc} shows the Top-1 and Top-5 accuracy trends for training ViT to various sparsity targets with different sparsification techniques. We observe generally, \mdgf and \sdgf are better than SR-STE, especially for high-sparsity targets.

\subsubsection{Image Classification \texorpdfstring{$\rightarrow$}{} Swin Transformer V2 (\xx{SwinV2})}
We train the SwinV2-Base model on imagenet-1k with hyperparameters presented in \autoref{table:swin_hyper}. We follow the hyperparameter setting in \citep{liu2022swin} for all SwinV2 experiments.
\begin{table}[!h]\centering
\caption{Hyperparameters used for training SwinV2 on ImageNet-1K.}
\begin{tabular}{lrr}\toprule
Batch Size & 128 \\
Training Epoches & 350 \\

Learning Rate & 1e-3 \\
LR Warmup Epoches & 20 \\
LR Decay schedular & Cosine \\

Decay Rate & 0.1 \\
Decay Epoches & 30 \\
Optimizer & AdamW  \\ 
Optimizer coefs &   beta1 = 0.9, beta2 = 0.999 \\
\bottomrule
\end{tabular}
\label{table:swin_hyper}
\end{table}

The detailed model configuration is the same as present in the original Microsoft research GitHub repo, \href{https://github.com/microsoft/Swin-Transformer/blob/main/configs/swinv2/swinv2_base_patch4_window16_256.yaml}{\text{SwinV2-base.yaml}}
The detailed list of all hyperparameters was taken from \href{https://github.com/microsoft/Swin-Transformer/blob/d19503d7fbed704792a5e5a3a5ee36f9357d26c1/config.py}{config.yaml}. For SwinV2-Base, the training phase takes $\approx$ 54 hours on 16 - A100  GPUs.

\subsubsection{Language Understanding \texorpdfstring{$\rightarrow$}{} \xx{T5X}}
We train the T5X-Base model on GLUE dataset with  hyperparameters presented in \autoref{table:t5x_hyper}. We follow the hyperparameter setting in \citep{t5_model} for all T5X training experiments.

The detailed model configuration is the same as present in the original Google research GitHub repo, \href{https://github.com/google-research/t5x}{\text{T5X model}}
T5X-Base's training phase takes $\approx$  22 hours on 8$\times$Google Cloud TPUv3 cores.

\begin{table}[!h]\centering
\caption{Hyperparameters used for training T5X on GLUE.}
\begin{tabular}{lrr}\toprule
Batch Size & 128 \\
Training Steps & 100k \\

Learning Rate & 1e-3 \\
LR Warmup Steps & 1000 \\
LR Decay schedular & Constant \\

Optimizer & AdamW  \\ 
Optimizer coefs &   beta1 = 0.9, beta2 = 0.999 \\
\bottomrule
\end{tabular}
\label{table:t5x_hyper}
\end{table}

\subsubsection{Language Translation Model \texorpdfstring{$\rightarrow$}{} \xx{Enc-Dec}}
\label{sec:langugae_model_impl}
We train an encoder-decoder-based model on WMT-17 with hyperparameters presented in \autoref{table:setup}. The model is inspired by the attention paper \citep{vaswani2017attention}. We follow the hyperparameter setting in \citep{devlin2019bert} to train all models. The training phase takes $\approx$ 8 hours on 32 - Google Cloud TPU v3 cores.

% \TODO{move this table to appendix}
\begin{table}[!ht]\centering
\caption{Model configurations and hyperparameters for training model on WMT.}
\begin{tabular}{lrr}\toprule
Number of Encoder Layers &6 \\
Number of Decoder Layer &6 \\
Hidden Dimension Size &1024 \\
Feed-Forward Dimension Size &4096 \\
Number of Attention Heads &16 \\
Max Sequence Length & 256 \\
Training Dataset & WMT-17 \\
Testing Dataset & WMT-14 \\

Batch Size &512 \\
Training Steps &200K \\

Learning Rate &0.0625 \\
LR Warmup Steps &1000 \\
Decay Factor &0.5 \\

Optimizer & Adam \\
Optimizer coefs & beta1 = 0.9, beta2 = 0.92 \\
\bottomrule
\end{tabular}
\label{table:setup}
\end{table}

% \section{Additional Results on the Impact of Sparsification Methods on Gradients}

% \ay{Add learning rate and other configs for Tiny ViT}

% \ay{Add the sensitivity for different layers}

\section{Codebase}
\label{sec:code_base}

Our ViT and SWINV2 codebase is made by modifying the TIMM code base of hugging-face vision transformers \citep{rw2019timm}. 
We add sparsity layers to various models and modify the training loop to support training recipes presented in this work.
Similarly, we modify the jax-based codebases for T5X and Language translation model experiments.
The source code is available at \href{https://github.com/abhibambhaniya/progressive_gradient_flow_nm_sparsity}{GitHub}.

\clearpage

\section{FLOPS Calculation}
\label{sec:flops_calc}

\begin{figure}[h]
    \centering
    \includegraphics[width=1\linewidth]{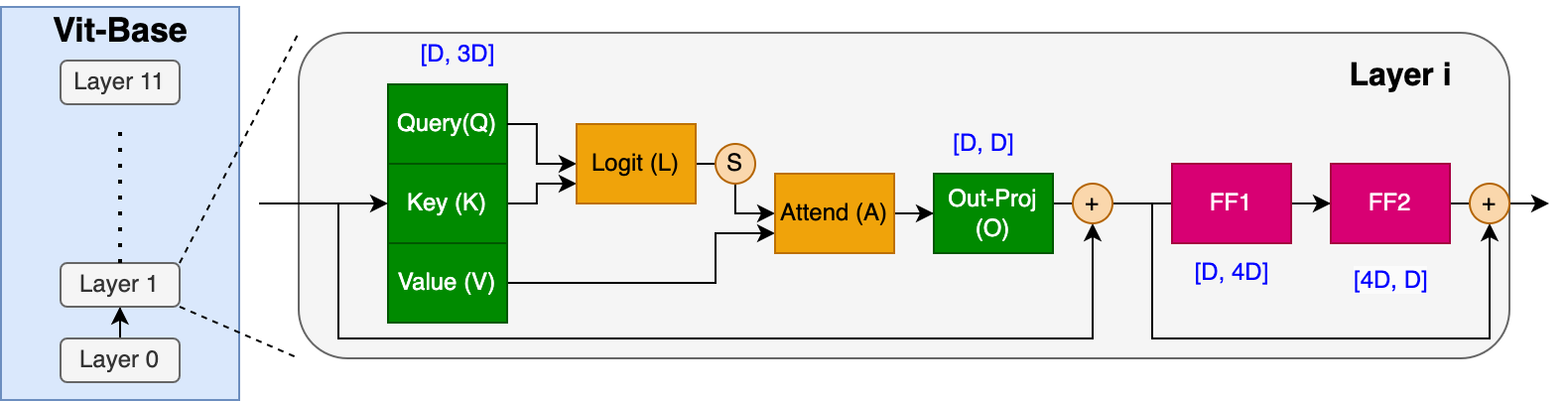}
    \caption{Operations for ViT base model. For sake of brevity, we only include the operators that take significant runtime. Parameter dimensions are mentioned in blue text near the corresponding operators.}
    \label{fig:vit_layers}
\end{figure}

\autoref{fig:vit_layers} shows various operators in ViT base model. The breakdown of flops, \autoref{tab:flops_breakdown}, shows that FF accounts for majority of the FLOPS and thus would be our main avenue of sparsification.

\begin{table}[!h]
    \centering
    \begin{tabular}{|c|c|c|c|}
        \toprule
        FLOPS (G) & Q/K/V/O & L/A  & FF1/FF2  \\
        \midrule
        Dense & 2.77  & 0.7 & 11.1 \\
        \bottomrule
    \end{tabular}
    \caption{Operator wise FLOPS breakdown for ViT-base.}
    \label{tab:flops_breakdown}
\end{table}

We calculate the total number of flops for the model as follows.

\begin{align*}
    FLOPS_{tot} &= FLOPS_{SA} + FLOPS_{FF} * S_{FF} \\
    FLOPS_{SA} &= FLOPS_{Q} + FLOPS_{K} + FLOPS_{V}+ FLOPS_{L} + FLOPS_{A}+ FLOPS_{O} \\
    FLOPS_{FF} &= FLOPS_{FF1} + FLOPS_{FF2}\\
\end{align*}

$FLOPS_{SA}$ is number of flops in self-attention layers which consists of QKV generation, 2 einsums (Logit and Attend) and output projection(O). 

$FLOPS_{FF}$ is number of flops of the 2 feed-forward layers.

Using these equations, We list the total FLOPS of ViT-base for various sparsity targets in \autoref{tab:all_flops_in_vit }.

\begin{table}[h]
    \centering
    \begin{tabular}{|c|c|c|c|}
    \toprule
    Sparsity  : $S_{FF}$   & $FLOPS_{SA}$ & $FLOPS_{FF}$ & $FLOPS_{tot}$ \\
    \midrule
        Dense : 1.0 & 12.51 & 22.19 & 34.71 \\
        2:4 (FF) : 0.5 &  12.51 &	11.1 & 23.61 \\
        1:4 (FF) : 0.25&  12.51 &	5.55 & 18.06 \\
        1:8 (FF) : 0.125&  12.51 & 2.77 & 	15.29 \\
        1:16 (FF) : 0.0625  & 12.51 &	1.39 &	13.90 \\
        1:32 (FF) : 0.03125 & 12.51 &	0.69 &	13.20 \\
        1:128 (FF) : 0.0078125 &12.51 &	0.17 &	12.69 \\
    \bottomrule
    \end{tabular}
    \caption{FLOPS(G) calculation for various level of sparsity in ViT-Base.}
    \label{tab:all_flops_in_vit }
\end{table}

\end{document}